\documentclass{article}

\usepackage[eandd,final]{neurips_2026}
\usepackage{graphicx}
\usepackage{amsmath} 
\usepackage[table]{xcolor}
\usepackage{makecell}
\usepackage{subcaption}
\usepackage{multicol}
\usepackage{multirow}
\usepackage{pgfplots}
\pgfplotsset{compat=1.18} % 推荐设置兼容版本（1.18 或您当前 LaTeX 环境支持的版本）

\usepackage{subcaption}

\usepackage{xcolor}

\usepackage{tikz}

\usepackage[utf8]{inputenc} % allow utf-8 input
\usepackage[T1]{fontenc}    % use 8-bit T1 fonts
\usepackage{hyperref}       % hyperlinks
\usepackage{url}            % simple URL typesetting
\usepackage{booktabs}       % professional-quality tables
\usepackage{amsfonts}       % blackboard math symbols
\usepackage{nicefrac}       % compact symbols for 1/2, etc.
\usepackage{microtype}      % microtypography
\usepackage{xcolor}         % colors
\usepackage{colortbl}
\usepackage[table]{xcolor}
\usepackage{listings}
\usepackage[most]{tcolorbox}

\newtcolorbox{promptbox}[1][]{
  colback=gray!5!white,    % 极浅的灰色背景
  colframe=gray!60!black,  % 深灰色边框
  fonttitle=\bfseries,     % 标题加粗
  title=#1,                % 允许传入标题
  boxrule=0.5mm,           % 边框粗细
  arc=2mm,                 % 圆角大小
  left=3mm, right=3mm, top=2mm, bottom=2mm, % 内边距
  fontupper=\small\sffamily % 使用无衬线字体，更符合现代排版
}

\title{Are Text-to-Image Models Inductivist Turkeys?\\A Counterfactual Benchmark for Causal Reasoning}

\author{%
  \textbf{Jiayi Lei}$^{1,2}$, 
  \textbf{Yuandong Pu}$^{1,2}$, 
  \textbf{Xingyu Han}$^{1}$, 
  \textbf{Rongpeng Zhu}$^{1}$, 
  \textbf{Jing Xu}$^{3}$, 
  \textbf{Jinyao Wang}$^{1}$ \\
  \textbf{Zijian Zhou}$^{1}$, 
  \textbf{Bin Fu}$^{2}$, 
  \textbf{Yuewen Cao}$^{2\dagger}$, 
  \textbf{Yihao Liu}$^{2\dagger}$, 
  \textbf{Hongsheng Li}$^{3\dagger}$ \\
  $^{1}$Shanghai Jiao Tong University, 
  $^{2}$Shanghai AI Laboratory, \\
  $^{3}$The Chinese University of Hong Kong \\
 $^{\dagger}$Corresponding author
}

\begin{document}

\maketitle
% \begin{figure}[h]
%     \centering
%     \includegraphics[width=\linewidth]{fig/fig1.pdf}
%     \caption{
%     We evaluate T2I models across both factual and counterfactual scenarios. While models such as FLUX.2-dev, Qwen-image, and Nano Banana perform reliably on standard factual prompts (L1), they exhibit a substantial decline when presented with explicit (L2) and implicit (L3) counterfactuals.
%     }
%     \label{fig:teaser}
% \end{figure}
% 1. 摘要部分
\begin{abstract}
    Text-to-image (T2I) generation models have achieved remarkable progress in producing visually realistic images from natural language prompts. Yet it remains unclear whether their success reflects genuine causal understanding or sophisticated pattern matching over visual-textual correlations. Inspired by Russell’s inductivist turkey, we introduce Counterfactual-World (CF-World), a counterfactual benchmark designed to investigate whether text-to-image models can generate images under rules that systematically contradict real-world priors. CF-World organizes each scenario into three progressive levels: factual generation under ordinary world knowledge, explicit counterfactual generation with direct visual instructions, and implicit counterfactual generation requiring causal deduction from altered rules. 
We evaluate open-source and closed-source text-to-image models using CF-Eval,
a scoring pipeline based on vision-language models. We report the scores at
each prompt level as the primary results and use Prior Resistance Rate (PRR)
and Reasoning Retention Rate (RRR) as auxiliary descriptive indicators.
All evaluated models score lower on counterfactual prompts than on factual
prompts. Diagnostic experiments reveal difficulties in following altered
rules and recombining rarely co-occurring visual attributes. The gains from
rephrasing object and attribute names are consistent with a role for learned
visual priors in counterfactual generation. Our benchmark and evaluation code are publicly available at \url{https://jylei16.github.io/CF-World.github.io/}
\end{abstract}

% 2. 正文各个 Section
\section{Introduction}

\begin{figure}[h]
    \centering
    \includegraphics[width=\linewidth]{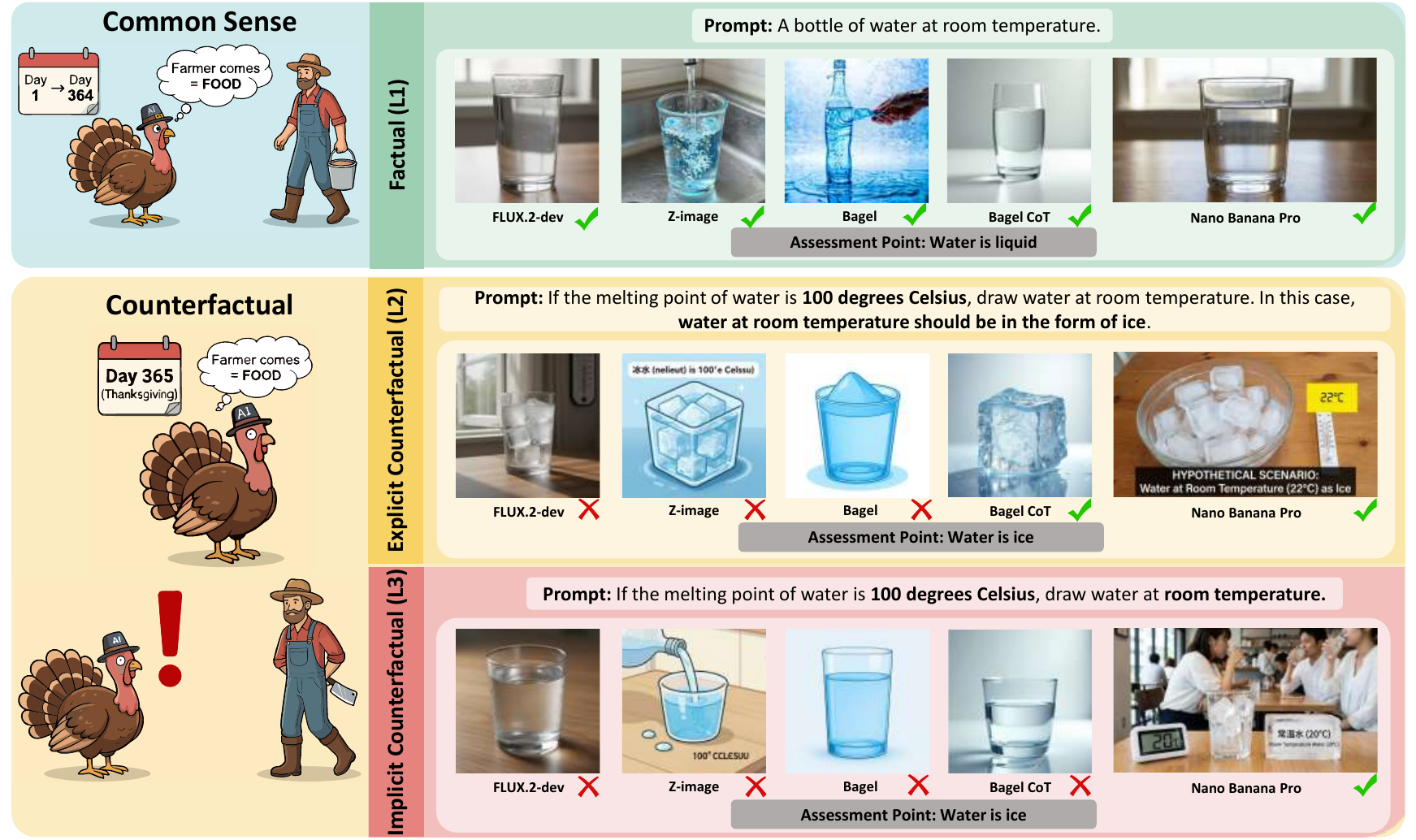}
    \caption{
    \textbf{Text-to-image models as inductivist turkeys.}
     Left: The inductivist turkey assumes food will always arrive based on past experience, failing to anticipate the counterfactual reality of Thanksgiving. Right: Current T2I models exhibit a similar flaw, evaluated through our three-level progressive framework. (1) \textbf{Factual ($L1$)}: The prompt aligns with real-world laws. (2) \textbf{Explicit Counterfactual ($L2$)}: Altered real-world laws with explicitly stated outcomes. (3) \textbf{Implicit Counterfactual ($L3$)}: Altered real-world laws without explicitly stated outcomes. Ultimately, while models perform reliably on $L1$, they exhibit a substantial decline on $L2$ and $L3$.
    }
    \label{fig:teaser}
\end{figure}

In Bertrand Russell's \textit{Inductivist Turkey} paradox, based on past experience, a turkey equates the farmer's arrival with food, only to be killed on Thanksgiving while awaiting its meal. According to Judea Pearl's Ladder of Causation, the turkey's failure stems from its cognition being restricted to the first level (Association), relying entirely on observed statistical correlations. It lacks the highest level of the ladder—\textit{counterfactual causal reasoning}—which requires the ability to mentally simulate alternative realities, such as the shift in the farmer's intent when Thanksgiving arrives.

Mirroring this inductivist turkey, current text-to-image (T2I) models excel on existing reasoning benchmarks, prompting widespread claims of reasoning capabilities. However, whether they truly possess counterfactual causal reasoning capabilities or merely remain confined to the association level like the turkey cannot be effectively evaluated using existing benchmarks. On one hand, recent reasoning-driven benchmarks by \citet{Niu2025WISEAW, Li2023ImageCG, Chen2025R2IBenchBR, Fu2024CommonsenseT2ICC, Li2025GIRBenchVB} focus primarily on conventional scenarios. This fails to separate causal reasoning from statistical priors, as successful generation could easily stem from retrieving high-frequency patterns memorized during training. On the other hand, existing counterfactual benchmarks \cite{Li2025ReplaceIT, zhao2024lost} typically target simple semantic combinations of unrelated objects, which are too superficial to assess a model's deductive reasoning and understanding of objective laws. Because of these evaluation gaps, a fundamental question remains unclear: have current T2I models actually climbed the causal ladder to achieve counterfactual causal reasoning?

We introduce the Counterfactual-World Benchmark (\textbf{CF-World}), designed to evaluate the counterfactual causal reasoning capabilities of T2I models. We propose a three-level progressive framework (Figure~\ref{fig:teaser}): it establishes a factual baseline ($L1$), introduces explicit counterfactual by providing both the altered objective laws and their outcomes ($L2$), and advances to implicit counterfactuals providing only the altered laws, requiring the model to deduce the visual outcomes ($L3$). This progressive design systematically isolates true logical deduction from mere statistical memorization. To rigorously quantify models’ performance, we develop \textbf{CF-Eval}, an automated pipeline assessing Visual Integrity, Assessment Point, and Logic Consistency. 
We report $L1$/$L2$/$L3$ scores as the primary results and use Prior Resistance Rate (PRR) and Reasoning Retention Rate (RRR) as auxiliary heuristic summaries of performance across levels. Under this rigorous setting, we surprisingly find that even state-of-the-art (SOTA) models experience a significant performance decline on $L2$ and $L3$, indicating their counterfactual causal reasoning capabilities remain relatively weak.

To understand this decline, we conduct diagnostic probes. \textbf{First, to isolate logical deduction}, we evaluate models on abstract symbolic elements under factual versus counterfactual rules. The universal performance drop in counterfactual scenarios suggests that, even without visual burdens, models fundamentally struggle with counterfactual causal reasoning. \textbf{Second, to isolate visual generation}, we test whether models can execute counterfactual visual recombination when causal reasoning is removed by combining rarely co-occurring concepts. We observe another universal performance drop. Digging deeper into this visual entanglement, we find that replacing high-frequency nouns with equivalent descriptive phrases yields notable improvements, revealing that models heavily rely on rigid text-image alignment shortcuts. \textbf{Ultimately, these dual failures highlight a fundamental bottleneck:} high-dimensional statistical priors constrain T2I models' decoupling abilities. Because they primarily learn pixel co-occurrences and text-image alignment, they struggle to decouple independent causal variables for logical reasoning and basic attribute modules for visual recombination, tending to default to the high-frequency commonsense priors found in their training data.

The main contributions of this paper are summarized as follows:

1) We introduce \textbf{CF-World}, the first counterfactual world knowledge benchmark structured through a systematic three-level progressive framework, bridging the critical gap in existing generative T2I evaluations by rigorously assessing both logical and causal reasoning under counterfactual premises.

2) We develop \textbf{CF-Eval}, an automated evaluation pipeline assessing visual integrity, assessment-point satisfaction, and logical consistency. We report level-specific performance together with auxiliary descriptive summaries (PRR and RRR) of cross-level performance.

3) Our \textbf{diagnostic experiments} reveal difficulties in following
altered rules and rendering rare concept pairs. Rephrasing improves
$L2$ scores, consistent with a role for visual priors.

\begin{table}[htbp]
\centering
\caption{\textbf{Comparison of CF-World with existing T2I evaluation benchmarks. Progressive design denotes matched factual, explicit-counterfactual, and implicit-counterfactual prompts for the same scenario, allowing separate examination of counterfactual instruction following and performance without explicit outcome descriptions.}}
\label{tab:comparison}
\begin{tabular}{lccc}
\toprule
\textbf{Benchmark} & \textbf{\makecell{Reasoning \\ Ability}} & \textbf{\makecell{Counterfactual \\ Setting}} & \textbf{\makecell{Progressive \\ Design}} \\ 
\midrule
WISE(~\cite{Niu2025WISEAW}) & \checkmark & $\times$ & $\times$ \\
VQAI(~\cite{Li2023ImageCG}) & \checkmark & $\times$ & $\times$ \\
R2I-Bench(~\cite{Chen2025R2IBenchBR}) & \checkmark & $\times$ & $\times$ \\
Commonsense-T2I(~\cite{Fu2024CommonsenseT2ICC}) & \checkmark & $\times$ & $\times$ \\
T2I-ReasonBench(~\cite{Sun2025T2IReasonBenchBR})  & \checkmark & $\times$ & $\times$ \\
GIR-Bench(~\cite{Li2025GIRBenchVB}) & \checkmark & $\times$ & $\times$ \\
\midrule
ELNP(~\cite{Li2025ReplaceIT}) & $\times$ & \checkmark & $\times$ \\
LC-Mis(~\cite{zhao2024lost}) & $\times$ & \checkmark & $\times$ \\
\midrule
\rowcolor{gray!10} \textbf{CF-World (Ours)} & \textbf{\checkmark} & \textbf{\checkmark} & \textbf{\checkmark} \\ 
\bottomrule
\end{tabular}
\end{table}
\section{Related Work}
\subsection{Text-to-Image Generation}

Text-to-image (T2I) synthesis has evolved rapidly, driven by breakthroughs across diverse architectural paradigms. Prominent approaches include diffusion-based methods \cite{Esser2024ScalingRF, Xie2025SANA1E, Qin2025LuminaImage2A, Yang2024ImprovingDI}, autoregressive models \cite{Sun2024AutoregressiveMB, Zhang2024VARCLIPTG, Chen2025JanusProUM, Wang2024Emu3NP}, and unified multimodal frameworks \cite{Xiao2024OmniGenUI, Xie2024ShowoOS, Zhou2024TransfusionPT, Tong2024MetaMorphMU, Sun2023GenerativePI}. To tackle increasingly complex user prompts, recent models have integrated advanced techniques such as Chain-of-Thought (CoT) prompting \cite{Liao2025ImageGenCoTET} and reinforcement learning \cite{Guo2025CanWG, Jiang2025T2IR1RI}. 

While these models excel at general generation, their reasoning capabilities face significant challenges when extended to counterfactual scenarios. In T2I synthesis, counterfactual generation requires rendering scenes that deviate from reality while preserving internal causal consistency. Foundational techniques attempt to achieve this by disentangling causal features via Generative Causal Models \cite{Yue2021CounterfactualZA} or employing fine-tuning strategies like DreamBooth \cite{Ruiz2022DreamBoothFT} to maintain subject identity across novel contexts \cite{He2023ADP}. Despite these capabilities, models often revert to training-data biases when faced with unusual concept combinations, leading to latent concept misalignment \cite{zhao2024lost}. Although recent advancements have attempted to correct these misalignments in physically implausible scenes through step-by-step latent space manipulation \cite{Li2025ReplaceIT}, these solutions remain constrained to visual attribute editing and concept co-occurrence. In contrast to these visually driven alignment methods, our work extends this paradigm by evaluating how models handle the systematic alteration of objective laws and causal logic.

\subsection{Text-to-Image Evaluation Benchmarks and Metrics}
To evaluate T2I generation, numerous benchmarks and metrics have been proposed in recent years. In terms of benchmarks, datasets such as GeckoNum(~\cite{Kajic2024EvaluatingNR}), Winoground(~\cite{Thrush2022WinogroundPV}), GenEval(~\cite{Ghosh2023GenEvalAO}), and GenAI-Bench(~\cite{Li2024GenAIBenchEA}) are widely used to assess compositional and numerical alignment. Meanwhile, OK-VQA(~\cite{Marino2019OKVQAAV}), WISE(~\cite{Niu2025WISEAW}), Commonsense-T2I(~\cite{Fu2024CommonsenseT2ICC}), R2I-Bench(~\cite{Chen2025R2IBenchBR}), T2I-ReasonBench~\cite{Sun2025T2IReasonBenchBR}, StructBench(~\cite{Zhuo2025FactualityMW}) and PICABench(~\cite{Pu2025PICABenchHF}) have been introduced to explore knowledge-based and commonsense generation. To quantify these abilities, various metrics have been developed, ranging from traditional visual-text alignment scores such as CLIPScore(~\cite{Hessel2021CLIPScoreAR}), DSGScore(~\cite{Cho2023DavidsonianSG}), and VQAScore(~\cite{Lin2024EvaluatingTG}), to LLM-assisted evaluators including LLMScore(~\cite{Lu2023LLMScoreUT}), SemVarEffect(~\cite{Zhu2024EvaluatingSV}), RIScore(~\cite{Zhao2025EnvisioningBT}), and WIScore(~\cite{Niu2025WISEAW}).
\section{Counterfactual-World (CF-World)}

\subsection{Overview}
\begin{figure}[t] 
    \centering
    % Subfigure (a): Data Distribution Pie Chart
    \begin{subfigure}[b]{0.30\textwidth}
        \centering
        \includegraphics[width=\textwidth]{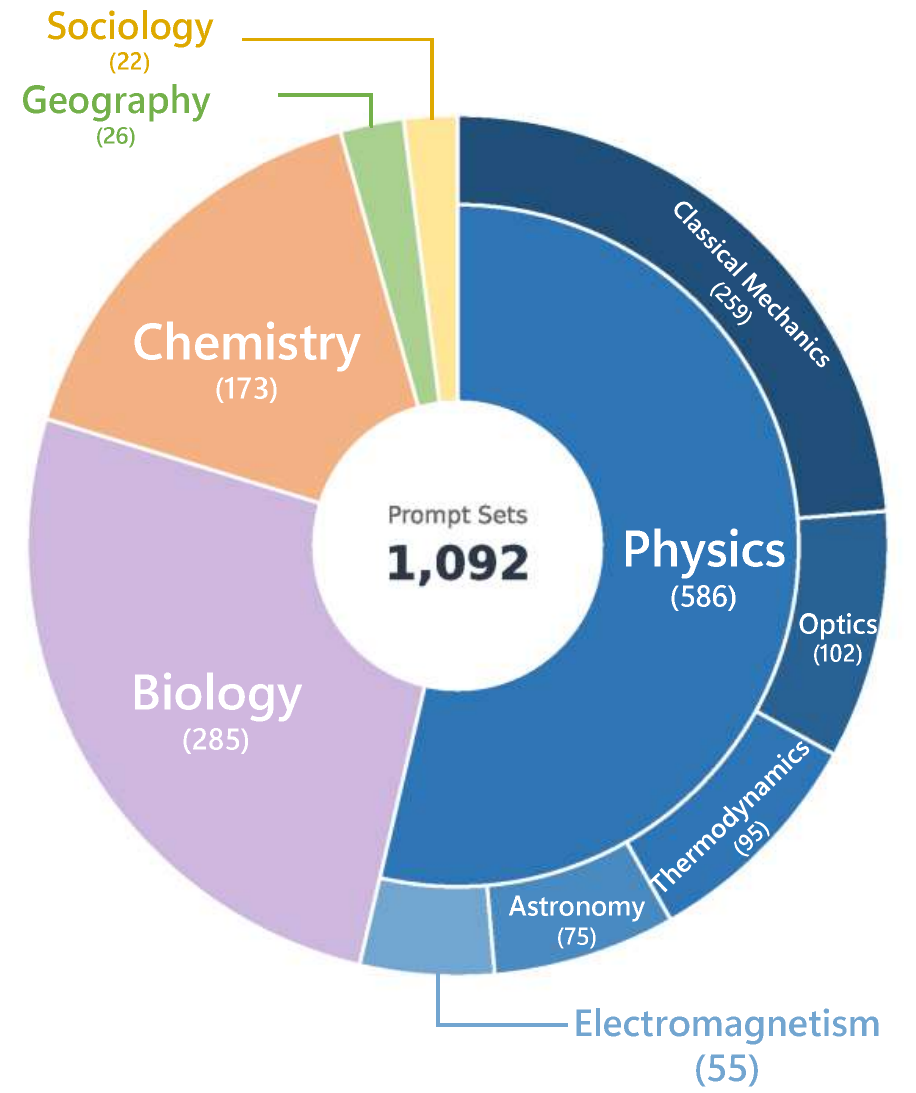}
        \caption{Data Distribution of CF-World}
        \label{fig:sub_pie}
    \end{subfigure}
    \hfill 
    % Subfigure (b): Overview Pipeline
    \begin{subfigure}[b]{0.69\textwidth}
        \centering
        \includegraphics[width=\textwidth]{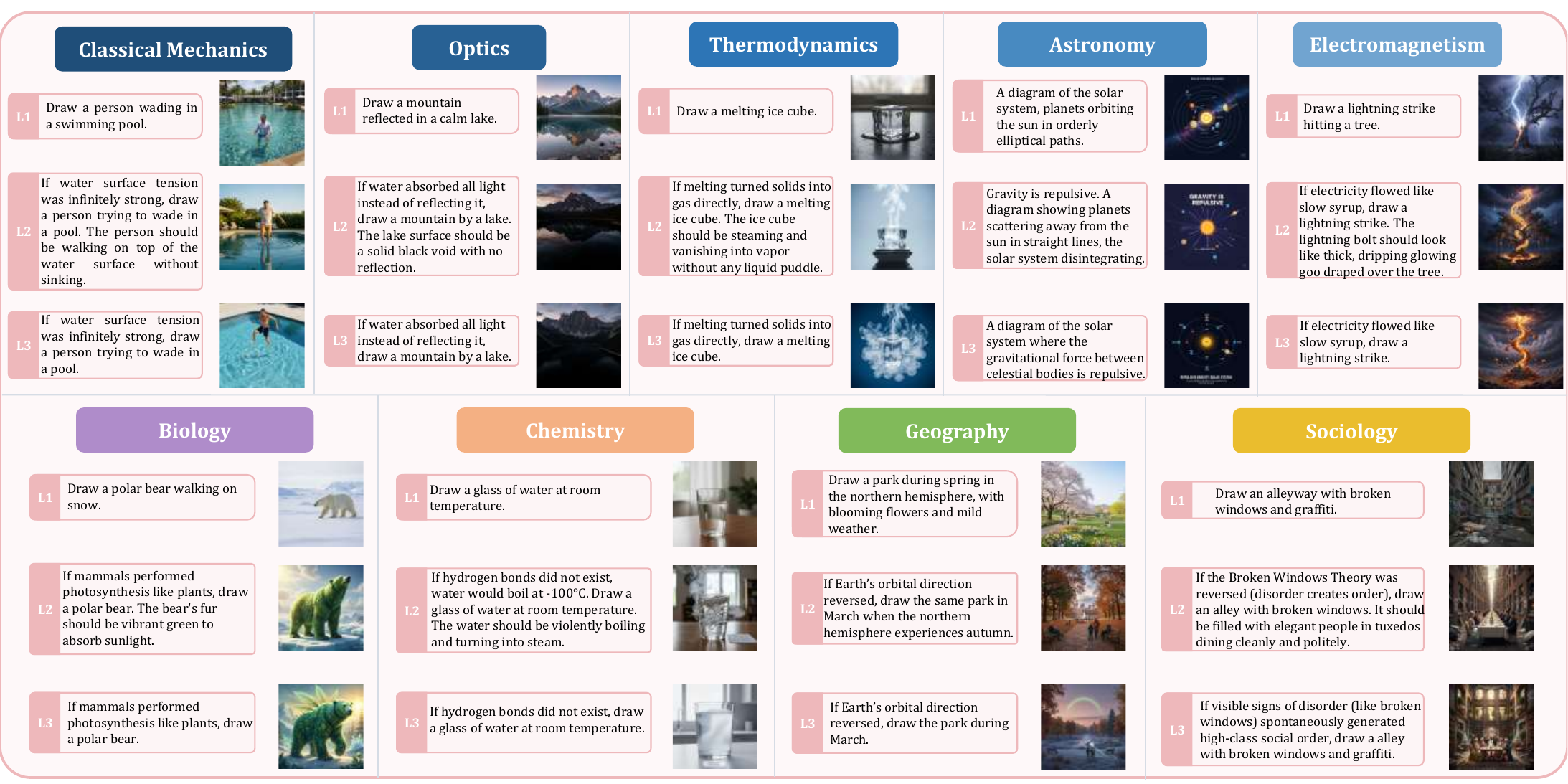}
        \caption{Overview of CF-World}
        \label{fig:sub_overview}
    \end{subfigure}
    
    \caption{The dataset distribution and selected qualitative examples of CF-World. (a) The data distribution of CF-World. (b) Qualitative examples across the five disciplines.}
    \label{fig:dataset_and_overview}
\end{figure}

To systematically evaluate the counterfactual causal reasoning capability of T2I models, we introduce Counterfactual-World (CF-World). As illustrated in Figure~\ref{fig:dataset_and_overview}, CF-World is a comprehensive benchmark comprising 1,092 groups and a total of 3,276 prompts. These prompts span five major disciplines: Physics (including branches such as Classical Mechanics, Optics, Thermodynamics, Astronomy, and Electromagnetism), Biology, Chemistry, Geography, and Sociology.

To isolate a model's reasoning capacity from its basic rendering, we design a three-level progressive framework for our prompts:
\textbf{Factual ($L1$):} Follows real-world objective laws to verify the model's basic priors.
\textbf{Explicit Counterfactual ($L2$):} Alters real-world laws but explicitly states the visual outcomes. This tests whether the model can overcome commonsense to render the counterfactual state.
\textbf{Implicit Counterfactual ($L3$):} Alters real-world laws without explicitly stating the outcomes, forcing the model to perform autonomous causal deduction. Each prompt is paired with a tailored assessment point, serving as the decisive visual criterion for automated scoring.

\subsection{Benchmark Construction}
To ensure the high quality and scientific validity of CF-World, we develop a rigorous pipeline encompassing taxonomy definition, data generation, and human-in-the-loop quality assurance.

\textbf{Taxonomy.}
CF-World groups counterfactual scenarios into five disciplines:
Physics, Biology, Chemistry, Geography, and Sociology. We select
laws commonly taught in middle-school curricula to define
the counterfactual worlds. Controlling the required domain knowledge
helps focus evaluation on understanding and applying altered rules
while reducing the influence of specialized expertise.

\textbf{Data Generation Pipeline.}
We manually curated textbook principles in physics, chemistry, biology,
geography, and sociology as seed knowledge. Using a fixed template,
Gemini-3-Pro generated candidate scenarios at three levels. L1 follows
real-world laws; L2 modifies a law and states the expected visual outcome;
L3 provides the modified law alone, requiring the T2I model to infer and
render the outcome. Gemini-3-Pro also generated an assessment point for
each prompt as an evaluation reference. The generation template is
provided in Appendix~\ref{app:data_generation_prompt}.

Generation followed four criteria. \textbf{Visual Unambiguity} requires
clear outcomes assessable by a VLM. The \textbf{Logical Deduction
Requirement} calls for reasoning beyond stylistic changes.
\textbf{Safety} excludes NSFW and body-horror content.
\textbf{Scientific Validity} requires logically sound deductions and
comprehensible visual targets under the stated premises.

\textbf{Human Review.}
Five co-authors reviewed 3,000 candidate scenario groups against the four criteria above.
A group was retained if at least two annotators approved it.
The review yielded 1,092 scenario groups, comprising 3,276 prompts, with a retention rate of 36.4\%.
The main reasons for rejection were insufficient visual unambiguity and problems with the required logical deduction.
Further details on the collection and review process are provided in Appendix~\ref{Collection_Process}.

\subsection{CF-Eval}
\begin{figure}[t]
    \centering
    \includegraphics[width=\textwidth]{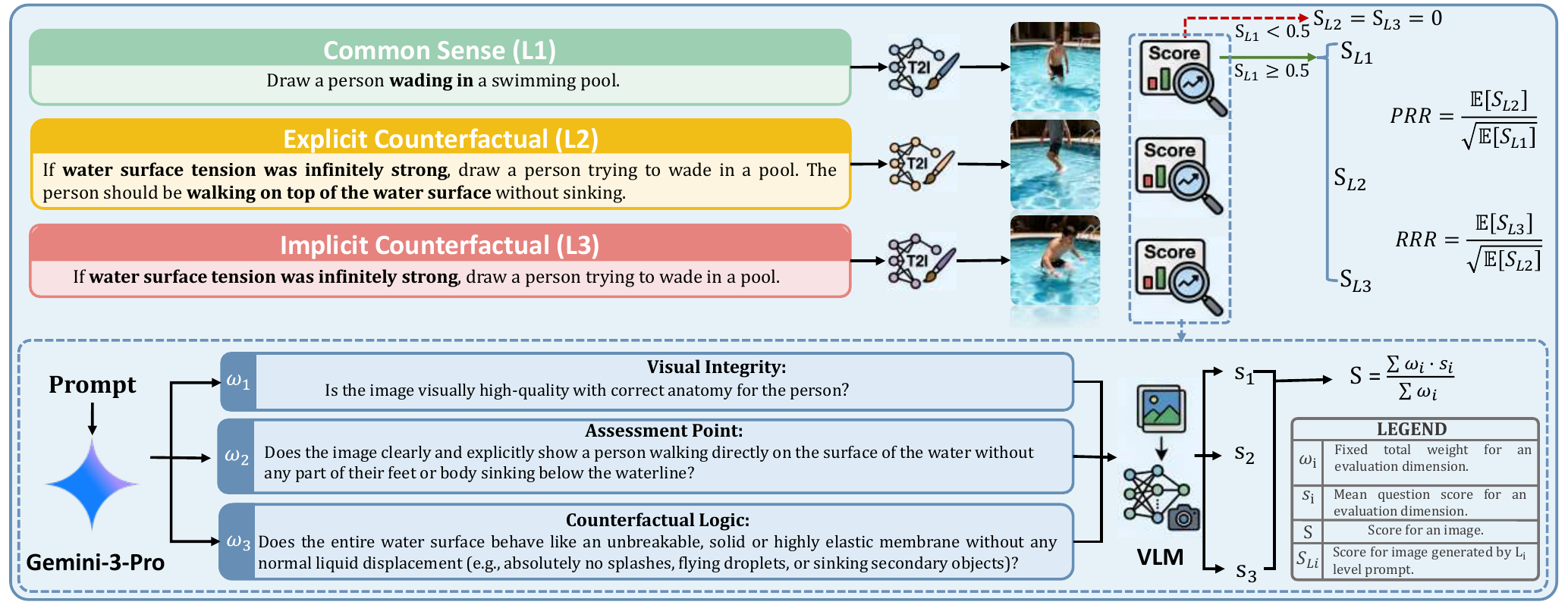}
    \caption{\textbf{CF-Eval.} 
   CF-Eval is a multi-dimensional scoring pipeline, featuring sequential thresholding ($S_{L1} \ge 0.5$) and two metrics: Prior Resistance Rate (PRR) and Reasoning Retention Rate (RRR).}
    \label{fig:pipeline}
\end{figure}
To evaluate the generative capabilities of models in factual and counterfactual scenarios, we propose \textbf{CF-Eval}. As illustrated in Figure~\ref{fig:pipeline}, CF-Eval is an automated evaluation pipeline driven by Vision-Language Models (VLMs). We detail its scoring mechanisms and metrics below.

\subsubsection{Evaluation Dimensions}
For each prompt, evaluation questions are generated by
Gemini-3-Pro across three dimensions. Each question is assigned
a weight to reflect its contribution to the evaluation:

\textbf{1) Visual Integrity (total weight: 1--3):}
Evaluates fundamental, style-agnostic image quality as a basic
viability threshold.

\textbf{2) Assessment Point (total weight: 12--16):}
Evaluates whether the main subject adheres to the factual or
counterfactual rule specified by the assessment point.

\textbf{3) Logic Consistency (total weight: 7--9):}
Verifies that the surrounding context aligns with the established
physical or counterfactual setting, ensuring global coherence.

To ensure consistent dimension-level weighting across L1, L2, and L3, we assign a fixed total weight to each dimension. When a dimension contains multiple evaluation questions, its total weight is distributed equally among those questions.

\subsubsection{Score Calculation and Thresholding}

For each generated image, the vision-language model (VLM) assigns
each evaluation question a score between 0 and 1. Let
$s_{l,d,j} \in [0,1]$ and $w_{l,d,j}$ denote the score and weight
of the $j$-th question in dimension $d$ at level $l$, respectively.
Let $n_{l,d}$ denote the number of questions in that dimension.
The raw score $B_l$ is calculated as the weighted average of all
question scores:

\begin{equation}
B_l =
\frac{
    \sum_{d \in \{\mathrm{VI}, \mathrm{AP}, \mathrm{LC}\}}
    \sum_{j=1}^{n_{l,d}} w_{l,d,j} s_{l,d,j}
}{
    \sum_{d \in \{\mathrm{VI}, \mathrm{AP}, \mathrm{LC}\}}
    \sum_{j=1}^{n_{l,d}} w_{l,d,j}
},
\end{equation}

\noindent where VI, AP, and LC denote Visual Integrity,
Assessment Point, and Logic Consistency, respectively.

We then apply a conditional threshold to the raw scores. For a
given scenario, the final L2 and L3 scores are retained only if
the raw L1 score meets the factual-generation threshold of 0.5;
otherwise, they are set to zero. This sequential design conditions the counterfactual scores on
the model meeting a factual-generation threshold for the
corresponding scenario. It reduces the risk of treating apparent
success on a counterfactual task as meaningful when the model
fails the corresponding factual-generation task. The threshold
of 0.5 was empirically calibrated based on alignment with human
judgments, as detailed in Appendix~\ref{threshold}.

\subsubsection{Auxiliary Descriptive Indicators: PRR and RRR}

We use the mean L1, L2, and L3 scores as the primary evaluation results. To provide complementary summaries of performance across levels, we report two auxiliary descriptive indicators: Prior Resistance Rate (PRR) and Reasoning Retention Rate (RRR):
\[
    PRR = \frac{\mathbb{E}[S_{L2}]}{\sqrt{\mathbb{E}[S_{L1}]}},
    \qquad
    RRR = \frac{\mathbb{E}[S_{L3}]}{\sqrt{\mathbb{E}[S_{L2}]}}.
\]
For positive preceding-level mean score $a$ and subsequent-level mean score $b$, both indicators take the form
$g(a,b) = \frac{b}{\sqrt{a}}
    = \sqrt{b \cdot \frac{b}{a}}.$
This heuristic combines the subsequent-level absolute score with its ratio to the preceding-level score. Compared with the ratio $b/a$ alone, it downweights cases with low subsequent-level performance. It also distinguishes models with similar subsequent-level scores but different preceding-level baselines. PRR summarizes explicit-counterfactual performance relative to factual performance, while RRR summarizes implicit-counterfactual performance relative to explicit-counterfactual performance.

\section{Experiments}
\begin{figure}[t]
    \centering
    \includegraphics[width=\textwidth]{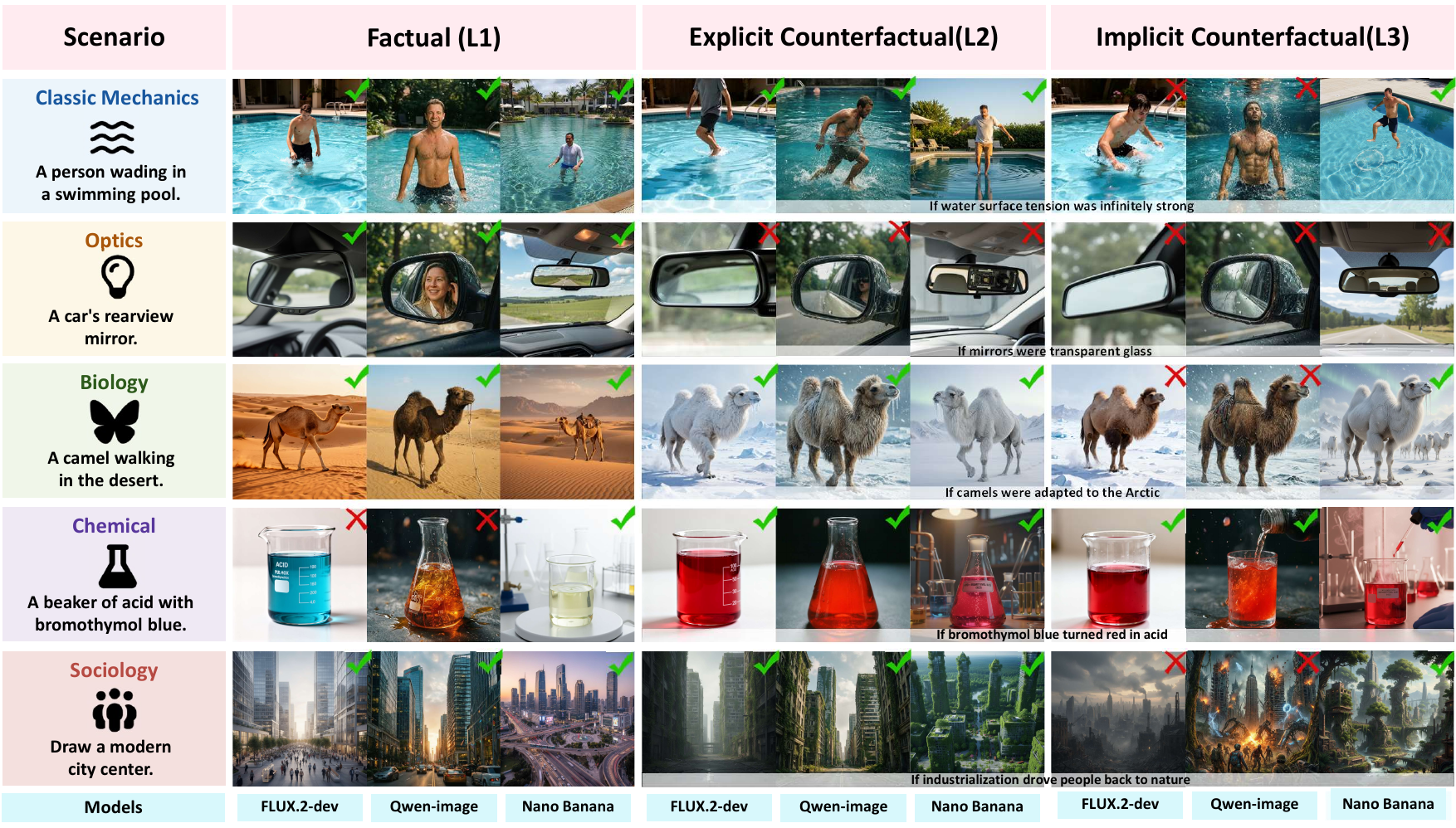} 
    \caption{\textbf{Qualitative Comparison of Model Generations.} A detailed visual comparison of selected models given identical prompts sampled from the five different scientific domains.}
    \label{fig:qualitative_comp}
\end{figure}

\subsection{Setup}
To evaluate the counterfactual reasoning capabilities of current text-to-image models, we select diverse state-of-the-art systems, as shown in Figure~\ref{fig:qualitative_comp}. Our evaluation suite encompasses prominent open-source models: SANA 1.5~\cite{Xie2025SANA1E}, Janus-Pro-7B~\cite{Chen2025JanusProUM}, Show-o2~\cite{Xie2025Showo2IN}, Z-image~\cite{Team2025ZImageAE}, Lumina-DiMOO~\cite{Xin2025LuminaDiMOOAO}, BAGEL and BAGEL-CoT~\cite{Deng2025EmergingPI}, OmniGen2~\cite{Wu2025OmniGen2TI}, FLUX.2-dev~\cite{bfl2025flux2}, and Qwen-Image. We also test leading closed-source models such as Nano Banana~\cite{nanobanana2025}, Nano Banana Pro, GPT-Image-1.5~\cite{openai2025b}, and Seedream 5.0~\cite{bytedance2025seedream}. To ensure scoring objectivity across these diverse architectures, we employ two highly capable Vision-Language Models (VLMs) as evaluators: Qwen3-VL-235B~\cite{qwen3technicalreport} and Gemini-3-Pro.

\subsection{Main Results and Analysis}
Table~\ref{tab:main_results} reports the primary scores for each evaluated model across three prompt levels: Factual ($L1$), Explicit Counterfactual ($L2$), and Implicit Counterfactual ($L3$), together with PRR and RRR as auxiliary descriptive indicators. We focus our analysis on level-specific performance and use the auxiliary indicators to contextualize cross-level changes.

\textbf{The Prior Lock-in in Counterfactual Generation.} 
Our systematic evaluation reveals a substantial performance decline from factual ($L1$) to explicit counterfactual generation ($L2$). While most open-source models achieve strong $L1$ scores, their $L2$ performance drops significantly, yielding PRRs largely below $0.50$. Interestingly, models with superior foundational capabilities (e.g., Qwen-Image) do not always exhibit proportional advantages in counterfactual tasks; they often yield lower PRRs than models with weaker $L1$ baselines. This paradox suggests that extensive reliance on training data exacerbates the "prior lock-in" effect, where visual representations and real-world knowledge become so deeply entangled that stronger priors actively hinder counterfactual rendering.

\textbf{Reasoning Improves Performance.}
Across the evaluated models, $L3$ scores are lower than $L2$ scores (see Figure~\ref{fig:degradation_bar_sub}), showing that counterfactual generation becomes more difficult when the expected visual outcome is no longer explicitly stated. BAGEL-CoT improves over BAGEL on $L3$ under both evaluators, indicating a benefit from the reasoning-enabled variant. Nevertheless, its $L3$ performance remains below its $L2$ performance. These observations motivate the targeted diagnostic experiments in Section~\ref{analysis}, which examine rule following and visual attribute recombination separately.

\textbf{Performance Comparison of Different Models} 
A clear performance gap exists between closed-source and open-source models. Top-tier closed-source models (e.g., Nano Banana Pro) maintain robust scores across both $L2$ and $L3$, which may be attributed to their use of large-scale, high-quality alignment data and specific architectural optimizations. Among open-weight models, architectural choices play a crucial role. Native multimodal and unified architectures (e.g., OmniGen2, Show-o2, Janus-Pro) are consistently outperformed by FLUX.2-dev. Despite the theoretical advantages of unified token spaces, our findings suggest that heavily scaled text encoders may currently be more effective at mitigating attribute entanglement than end-to-end unified architectures.

\begin{table*}[t]
    \centering
    \caption{Main evaluation results on CF-World. $L1$/$L2$/$L3$ scores are the primary results; PRR and RRR are auxiliary heuristic aggregates. The best open-source and closed-source results in each column are highlighted in \colorbox{cyan!15}{blue} and \colorbox{pink!30}{pink}, respectively. Ties after rounding to two decimal places are all highlighted.}
    \label{tab:main_results}
    \resizebox{\textwidth}{!}{
    \begin{tabular}{cccccc ccccc}
        \toprule
        \multirow{2}{*}{\textbf{Model}}
        & \multicolumn{5}{c}{\textbf{Qwen3-VL-235B}}
        & \multicolumn{5}{c}{\textbf{Gemini-3-Pro}} \\
        \cmidrule(lr){2-6} \cmidrule(lr){7-11}
        & \textbf{L1} & \textbf{L2} & \textbf{L3}
        & \textbf{PRR$\uparrow$} & \textbf{RRR$\uparrow$}
        & \textbf{L1} & \textbf{L2} & \textbf{L3}
        & \textbf{PRR$\uparrow$} & \textbf{RRR$\uparrow$} \\

        \midrule
        \multicolumn{11}{>{\columncolor{yellow!25}}c}
        {\textit{Open-Source Models}} \\
        \midrule

        SANA 1.5
        & 0.82 & 0.35 & 0.23 & 0.39 & 0.38
        & 0.75 & 0.30 & 0.17 & 0.35 & 0.31 \\

        Janus-Pro-7B
        & 0.80 & 0.30 & 0.21 & 0.33 & 0.39
        & 0.69 & 0.21 & 0.11 & 0.25 & 0.24 \\

        Show-o2
        & 0.76 & 0.31 & 0.19 & 0.35 & 0.35
        & 0.66 & 0.25 & 0.15 & 0.31 & 0.29 \\

        Z-image
        & 0.82 & 0.38 & 0.21 & 0.42 & 0.33
        & 0.75 & 0.34 & 0.16 & 0.39 & 0.27 \\

        Lumina-DiMOO
        & 0.74 & 0.31 & 0.19 & 0.36 & 0.34
        & 0.70 & 0.29 & 0.16 & 0.35 & 0.30 \\

        BAGEL
        & 0.80 & 0.29 & 0.16 & 0.33 & 0.30
        & 0.73 & 0.29 & 0.15 & 0.34 & 0.29 \\

        BAGEL-CoT
        & \colorbox{cyan!15}{0.88}
        & \colorbox{cyan!15}{0.44}
        & \colorbox{cyan!15}{0.30}
        & \colorbox{cyan!15}{0.47}
        & \colorbox{cyan!15}{0.45}
        & 0.82 & 0.42 & 0.26 & 0.46
        & \colorbox{cyan!15}{0.41} \\

        OmniGen2
        & 0.76 & 0.31 & 0.19 & 0.36 & 0.34
        & 0.70 & 0.29 & 0.18 & 0.35 & 0.33 \\

        FLUX.2-dev
        & 0.81 & 0.42 & 0.26
        & \colorbox{cyan!15}{0.47}
        & 0.40
        & \colorbox{cyan!15}{0.83}
        & \colorbox{cyan!15}{0.48}
        & \colorbox{cyan!15}{0.29}
        & \colorbox{cyan!15}{0.53}
        & \colorbox{cyan!15}{0.41} \\

        Qwen-Image
        & 0.83 & 0.35 & 0.24 & 0.39 & 0.40
        & 0.80 & 0.38 & 0.23 & 0.42 & 0.38 \\

        \midrule
        \multicolumn{11}{>{\columncolor{blue!15}}c}
        {\textit{Closed-Source Models}} \\
        \midrule

        Nano Banana
        & 0.93 & 0.65 & 0.55 & 0.67 & 0.68
        & 0.88 & 0.65 & 0.52 & 0.69 & 0.65 \\

        Nano Banana Pro
        & \colorbox{pink!30}{0.95}
        & \colorbox{pink!30}{0.67}
        & \colorbox{pink!30}{0.59}
        & \colorbox{pink!30}{0.69}
        & \colorbox{pink!30}{0.71}
        & \colorbox{pink!30}{0.93}
        & \colorbox{pink!30}{0.76}
        & \colorbox{pink!30}{0.67}
        & \colorbox{pink!30}{0.79}
        & \colorbox{pink!30}{0.76} \\

        GPT-Image-1.5
        & 0.92 & 0.66 & 0.50
        & \colorbox{pink!30}{0.69}
        & 0.61
        & 0.90 & 0.74 & 0.56 & 0.77 & 0.65 \\

        Seedream 5.0
        & 0.91 & 0.63 & 0.51 & 0.66 & 0.64
        & 0.89 & 0.73 & 0.61 & 0.77 & 0.71 \\

        \bottomrule
    \end{tabular}
    }
\end{table*}

\subsection{Human-VLM Scoring Consistency}
\label{sec:scoring_agreement}
\begin{figure}[t]
    \centering
    
    % --- 左侧子图 (a)：打分一致性分析 (占 30% 宽度) ---
    \begin{subfigure}{0.38\textwidth}
        \centering
        \includegraphics[width=\textwidth]{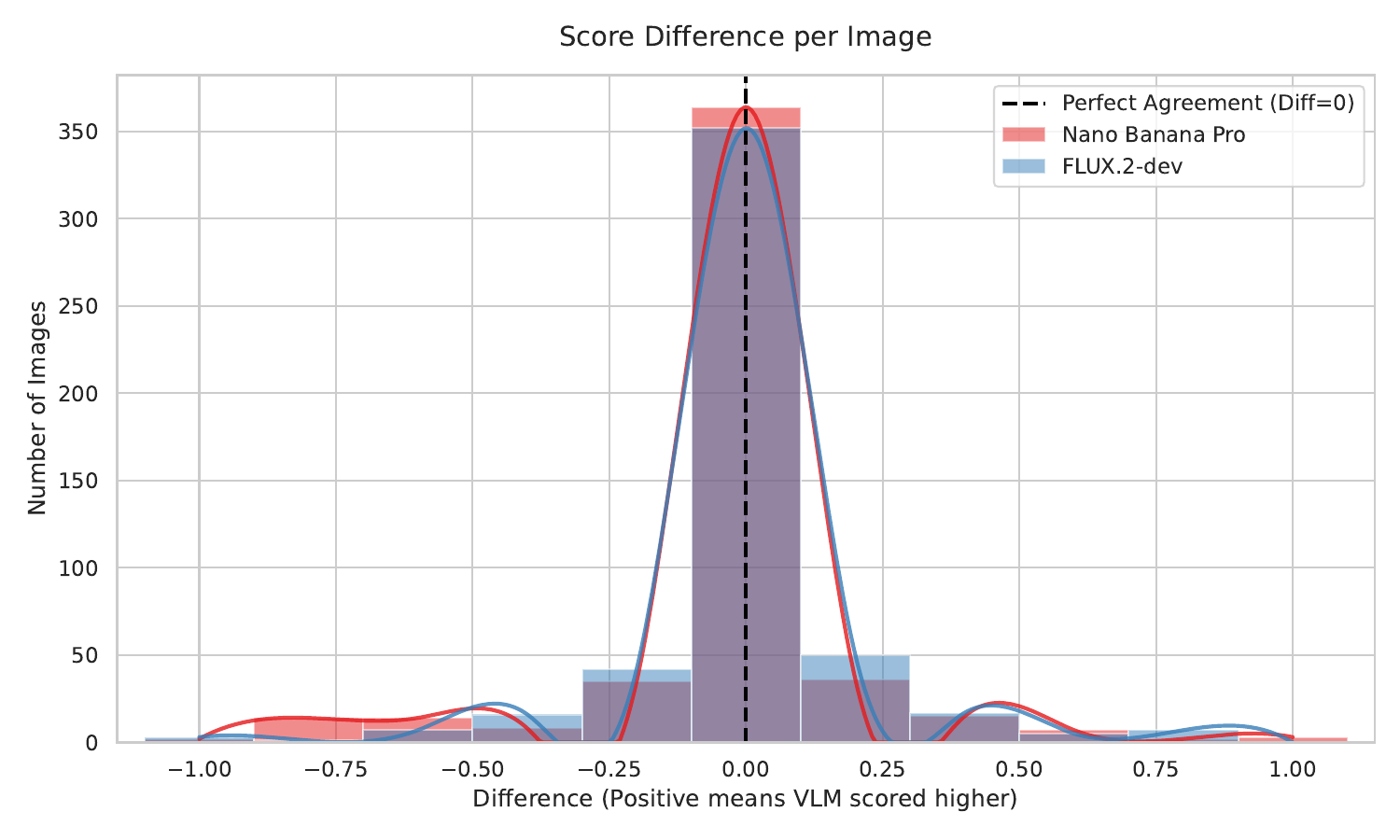}
        \caption{Human-VLM Consistency}
        \label{fig:human_eval_sub}
    \end{subfigure}
    \hfill % 自动填充中间空白
    % --- 右侧子图 (b)：性能断崖柱状图 (占 68% 宽度) ---
    \begin{subfigure}{0.60\textwidth}
        \centering
        % 替换为你保存的柱状图 PDF 路径
        \includegraphics[width=\textwidth]{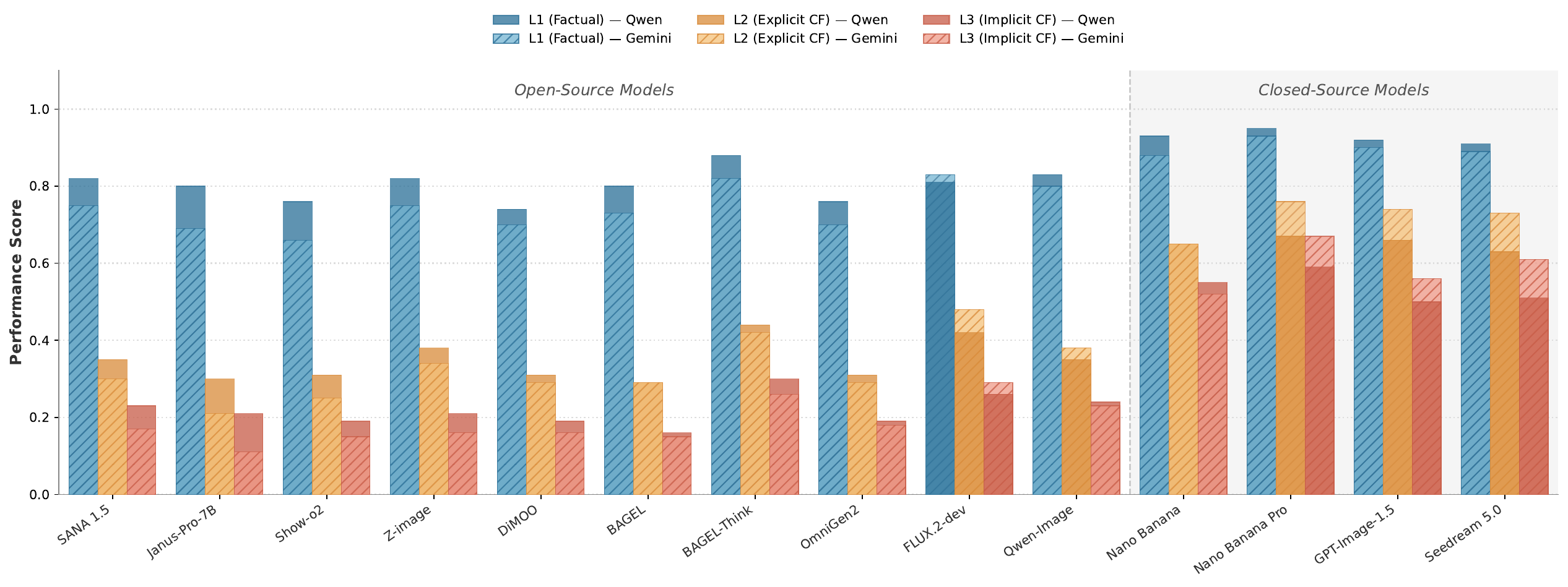} 
        \caption{Performance Degradation}
        \label{fig:degradation_bar_sub}
    \end{subfigure}
    
    % --- 全局图注 ---
    \caption{\textbf{Quantitative Evaluation and Consistency Analysis.} 
    \textbf{(a)} The distribution of score differences ($Score_{VLM} - Score_{Human}$) per image. The peak at $0$ indicates strong alignment between the VLM and humans. 
    \textbf{(b)} Performance degradation across factual ($L1$) and counterfactual ($L2, L3$) tasks. Open-source models exhibit a severe performance drop when transitioning to counterfactual scenarios, whereas closed-source models demonstrate stronger reasoning retention.}
    \label{fig:quant_analysis}
\end{figure}

\textbf{Human--VLM Agreement.}
We assessed agreement on 1,000 randomly sampled images, comprising 500 images from Nano Banana Pro and 500 from FLUX.2-dev. Two annotators---an AI PhD student and a physics undergraduate from the dataset-review team---independently evaluated all images using the prompt-specific questions generated by CF-Eval. They viewed the images and evaluation questions without access to the generating model's identity, VLM scores, or the other annotator's ratings. The mean of the two annotators' continuous scores served as the human reference.

Human and VLM scores were normalized to $[0,1]$ and aggregated using identical methods. As illustrated in Figure~\ref{fig:quant_analysis}(a), the differences between Gemini-3-Pro and human scores ($Score_{VLM} - Score_{Human}$) are predominantly concentrated within $[-0.125,0.125]$, indicating close alignment in continuous scores. For categorical agreement, scores were discretized into five ordered categories. Inter-annotator quadratic-weighted Cohen's $\kappa$ was 0.85. Gemini-3-Pro and Qwen3-VL achieved overall human--VLM agreement of $\kappa = 0.706$ and $\kappa = 0.760$, respectively. On $L2$/$L3$, agreement ranged from 0.577 to 0.698 across the two evaluators. Full procedures and results by prompt level and generating model are provided in Appendix~C.

\textbf{Inter-Judge Agreement.}
Across all 14 models, Spearman rank correlations between Gemini-3-Pro
and Qwen3-VL were 0.901, 0.938, and 0.850 for $L1$, $L2$, and $L3$,
respectively, and 0.925 and 0.849 for PRR and RRR. Item-level agreement
was $\kappa = 0.716$ (quadratic-weighted Cohen's $\kappa$), with 70.7\%
exact agreement. Under both judges, closed-source models outperformed
open-source models on $L2$ and $L3$, though ranks among closely performing
open-source models varied (Appendix~\ref{app:human_vlm_agreement}).

\section{Why Models Fail: A Decoupling Perspective}  
\label{analysis}
To investigate the cause of text-to-image models' deficiency in counterfactual causal reasoning, triggering degradation on $L2$ and $L3$ tasks, we design three targeted mechanistic experiments. The first two isolate two orthogonal failure axes: \textbf{logical reasoning} and \textbf{visual recombination}. The third, De-nominalization, serves as a diagnostic bridge that traces both failures to a shared lexical root.

\begin{table*}[t]
    \centering
    \caption{
        Comprehensive results of the mechanistic investigation.
        All raw metrics are scaled to 0--1.
        ``Fact.'' and ``CF'' denote factual and counterfactual settings,
        respectively.
        For de-nominalization, red superscripts indicate the performance
        gain over the original L2 prompts.
        The best and second-best raw results in each column are highlighted
        in \colorbox{green!15}{green} and \colorbox{yellow!20}{yellow},
        respectively.
    }
    \label{tab:mechanistic_results}
    \resizebox{0.9\textwidth}{!}{
    \begin{tabular}{c c cc|cc|cc}
        \toprule
        \multirow{2}{*}{\textbf{Model}} &
\multirow{2}{*}{\textbf{Type}} &
\multicolumn{2}{c|}{\textbf{Rule Decoupling}} &
\multicolumn{2}{c|}{\textbf{Attribute Decoupling}} &
\multicolumn{2}{c}{\textbf{De-nominalization}} \\
        \cmidrule(lr){3-4}
        \cmidrule(lr){5-6}
        \cmidrule(lr){7-8}
        & & \textbf{Fact.} & \textbf{CF}
          & \textbf{Fact.} & \textbf{CF}
          & \textbf{L2} & \textbf{De-norm} \\
        \midrule

        SANA 1.5
        & Diff.
        & 0.31 & 0.30
        & 0.94 & 0.83
        & 0.35
        & 0.37\textsuperscript{\textcolor{red}{+0.02}} \\

        Janus-Pro-7B
        & Unif.
        & 0.19 & 0.07
        & 0.97 & 0.83
        & 0.30
        & 0.31\textsuperscript{\textcolor{red}{+0.01}} \\

        Show-o2
        & Unif.
        & 0.39 & 0.37
        & 0.92 & 0.80
        & 0.31
        & 0.37\textsuperscript{\textcolor{red}{+0.06}} \\

        Z-image
        & Diff.
        & \colorbox{green!15}{0.61}
        & \colorbox{green!15}{0.53}
        & \colorbox{yellow!20}{0.98}
        & \colorbox{yellow!20}{0.89}
        & 0.38
        & 0.43\textsuperscript{\textcolor{red}{+0.05}} \\

        Lumina-DiMOO
        & Unif.
        & 0.38 & 0.34
        & 0.97 & 0.82
        & 0.31
        & 0.35\textsuperscript{\textcolor{red}{+0.04}} \\

        BAGEL
        & Unif.
        & 0.29 & 0.22
        & \colorbox{yellow!20}{0.98}
        & 0.83
        & 0.29
        & 0.31\textsuperscript{\textcolor{red}{+0.02}} \\

        BAGEL-CoT
        & Unif.
        & 0.38 & 0.32
        & 0.97
        & \colorbox{green!15}{0.90}
        & \colorbox{green!15}{0.44}
        & \colorbox{yellow!20}{0.46}
          \textsuperscript{\textcolor{red}{+0.02}} \\

        OmniGen2
        & Unif.
        & 0.33 & 0.25
        & 0.96 & 0.81
        & 0.31
        & 0.35\textsuperscript{\textcolor{red}{+0.04}} \\

        FLUX.2-dev
        & Diff.
        & \colorbox{yellow!20}{0.53}
        & \colorbox{yellow!20}{0.52}
        & \colorbox{green!15}{0.99}
        & \colorbox{green!15}{0.90}
        & \colorbox{yellow!20}{0.42}
        & \colorbox{green!15}{0.51}
          \textsuperscript{\textcolor{red}{+0.09}} \\

        Qwen-Image
        & Diff.
        & 0.40 & 0.40
        & 0.97 & 0.86
        & 0.35
        & 0.37\textsuperscript{\textcolor{red}{+0.02}} \\

        \bottomrule
    \end{tabular}
    }
\end{table*}

\subsection{Causal Decoupling}
To isolate logical rule execution from visual generation complexity, we
evaluate models on a curated symbolic benchmark of 198 prompts covering
33 objective rules. Each rule includes 1--2 factual baselines and 4--5
counterfactual variants with multi-dimensional perturbations, such as
changing gravity to point leftward, rightward, or upward. Each prompt
includes an assessment point specifying the intended rule condition,
which Qwen3-30B converts into a targeted evaluation question scored by
Qwen3-VL-235B. This design reduces the influence of visual quality and
focuses evaluation on counterfactual rule following.

As Table~\ref{tab:mechanistic_results} shows, most models obtain low
absolute scores under counterfactual rules, even with reduced visual
clutter and object recognition demands. This suggests difficulties in
grounded, compositional rule execution beyond complex visual rendering,
although performance drops vary substantially across models.
Factual and counterfactual rule execution are closely related:
models performing better on factual rules often achieve higher
counterfactual scores. This pattern suggests that failures may involve
general limitations in symbolic rule representation and application,
alongside interference from memorized factual priors. Diffusion-based
models tend to achieve higher factual and counterfactual scores than
unified models, with Z-image and FLUX.2-dev among the strongest performers.
This architectural trend warrants caution given the limited model sample.
Overall, some models preserve performance better under rule perturbations,
but robust counterfactual rule following remains broadly underdeveloped.

\subsection{Attribute Decoupling}
To examine whether image generation models can recombine visual attributes
beyond frequent co-occurrences with reduced rule-level reasoning demands,
we evaluate them in an attribute decoupling setting. We sample $113$ rare
concept pairs from LC-Mis~\cite{zhao2024lost} as the counterfactual condition.
Gemini-3-Pro constructs a common co-occurring counterpart for each pair
and converts both into paired factual and counterfactual prompts.
Qwen3-VL-235B evaluates the generated images on a normalized scale of
$[0,1]$, assessing whether the required concepts are present and their
relationship is correctly instantiated.

As Table~\ref{tab:mechanistic_results} shows, models perform strongly
on factual pairs but consistently score lower on rare pairs, indicating
limited ability to recombine concepts beyond frequent co-occurrences.
Together with the rule-following experiment, these results identify
difficulties in both visual recombination and rule execution. We interpret
scores within each experiment given their different test items and
scoring criteria.

\subsection{De-nominalization}
Building on the attribute decoupling analysis above, we further examine whether such object--attribute entanglement also affects L2 counterfactual generation. We therefore conduct a de-nominalization experiment on L2 prompts, replacing only the target object or attribute nouns in the inferred outcome with descriptive phrases while preserving the counterfactual law. This tests whether bypassing explicit nominal cues facilitates attribute decoupling by reducing default object--attribute associations.

As Table~\ref{tab:mechanistic_results} shows, de-nominalization consistently improves performance across all models, but the gains are generally modest. The largest improvement is observed for FLUX.2-dev ($+0.09$), followed by Z-image and Show-o2 ($+0.06$), while several models, including Janus-Pro-7B, SANA 1.5, and BAGEL-CoT, show only marginal gains. 
These results show that explicit object or attribute names introduce measurable interference by activating learned visual priors, as de-nominalization consistently improves performance across models. However, the limited magnitude of these gains suggests that lexical priors are not the primary source of $L2$ failures. Consistent with the attribute decoupling results, this finding further indicates that the main bottleneck of current models lies in reasoning over higher-level generative rules, rather than merely composing perceptual elements.

\subsection{Analysis and Discussion}
\textbf{1) Asymmetric Decoupling.} Factual attribute rendering approaches perfection ($\geq0.92$), yet factual rule execution remains poor ($\leq0.61$). This asymmetry is quantitative and qualitative: attribute scores retain over $0.81$ of their factual value under counterfactual stress, whereas rule scores collapse to $0.37$. We hypothesize that generative models master shallow visual interpolation, but lack physical grounding for logical deduction.

\textbf{2) Lexical Vulnerability.} Scaled text encoders (e.g., FLUX.2-dev) benefit from de-nominalization ($+0.09$), while unified models (e.g., Janus-Pro-7B) show minimal gains. This suggests that diffusion model entanglement reflects lexical shortcuts, whereas unified models exhibit semantic-level entanglement enforcing conceptual priors regardless of phrasing.

\textbf{3) Two Regimes of Entanglement.} These diagnostics reveal two failure regimes. Diffusion models suffer from shallow, lexical entanglement: their de-nominalization gains show that priors are bound to surface word embeddings and partially unlocked by rephrasing. Unified models suffer from deep, semantic entanglement: negligible gains indicate that conceptual priors persist regardless of phrasing, residing in the shared representation space. This distinction implies different remedies---prompt- or encoder-level intervention may suffice for diffusion models, but unified models require representation-level grounding to achieve genuine compositional reasoning.

\section{Conclusion and Limitations}
Extensive evaluations reveal that while SOTA generative models excel in factual settings, their performance declines significantly under counterfactual conditions. Our mechanistic investigation demonstrates that this degradation stems from a fundamental inability to decouple: models fail to decouple objective world knowledge from default scenarios and struggle to separate visual attributes from their corresponding subjects, remaining deeply entangled in high-frequency priors. More importantly, compared with failures in visual attribute recombination, the inability to decouple and reason over higher-level generative rules constitutes the dominant performance bottleneck.

While our current work is fundamentally diagnostic and does not propose an algorithmic solution to concept entanglement, identifying these root causes is a critical first step. We hope CF-World will serve as a robust testing ground, inspiring future research to develop novel decoupling mechanisms and advance multimodal models from prior-driven generation toward genuine causal reasoning.

% \begin{ack}
% We thank Shanghai Artificial Intelligence Laboratory
% for providing GPU resources and model API access
% for this research.

% The authors declare no competing interests.
% \end{ack}
% 4. 参考文献部分 (使用独立的 .bib 文件)
\bibliographystyle{plainnat} % 设置参考文献的排版风格
\bibliography{references}    % 告诉 LaTeX 你的引用文件叫 references.bib
\clearpage
% 5. 附录部分

\appendix
 \section{Datasheet for Datasets}
Following the standard practices for dataset documentation, we provide a comprehensive datasheet for the CF-World benchmark. 

\subsection{Motivation and Composition}
The CF-World dataset was created to systematically probe whether current Text-to-Image (T2I) models possess genuine causal understanding or merely rely on superficial visual-textual co-occurrences. The dataset consists of $N=1092$ unique counterfactual scenarios. Each scenario is expanded into three progressive levels (L1: Factual, L2: Explicit Counterfactual, L3: Implicit Counterfactual), resulting in a total of 3276 distinct prompts. 

\subsection{Collection Process and Maintenance}
\label{Collection_Process}
The initial prompts were generated using the Gemini-3-Pro model and subsequently subjected to human-in-the-loop filtering. The five annotators included a PhD student in artificial intelligence, a master's student in chemistry, and three undergraduate students majoring in physics, bioinformatics, and automation, respectively. All annotators were co-authors of this paper. The dataset was hosted on Hugging Face and maintained by the authors. A \texttt{croissant.json} file is included in the repository to ensure compliance with Responsible AI (RAI) metadata standards.

\section{Empirical Calibration of the Factual Threshold}
\label{threshold}

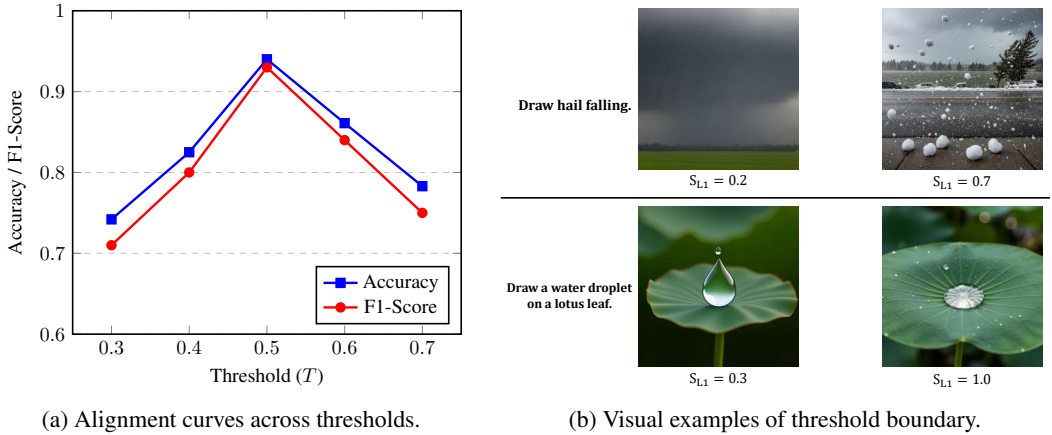
\begin{figure}[h]
\centering
% ==================== 左子图：折线图 ====================
\begin{subfigure}{0.45\textwidth}
\centering
\begin{tikzpicture}[scale=0.75]
\begin{axis}[
    xlabel={Threshold ($T$)},
    ylabel={Accuracy / F1-Score},
    xmin=0.25, xmax=0.75,
    ymin=0.6, ymax=1.0,
    xtick={0.3, 0.4, 0.5, 0.6, 0.7},
    ytick={0.6, 0.7, 0.8, 0.9, 1.0},
    legend pos=south east,
    ymajorgrids=true,
    grid style=dashed,
    thick
]
\addplot[
    color=blue,
    mark=square*,
    line width=1.2pt
]
coordinates {
    (0.3, 0.742) (0.4, 0.825) (0.5, 0.940) (0.6, 0.861) (0.7, 0.783)
};
\addlegendentry{Accuracy}

\addplot[
    color=red,
    mark=otimes*,
    line width=1.2pt
]
coordinates {
    (0.3, 0.71) (0.4, 0.80) (0.5, 0.93) (0.6, 0.84) (0.7, 0.75)
};
\addlegendentry{F1-Score}
\end{axis}
\end{tikzpicture}
\caption{Alignment curves across thresholds.}
\label{fig:threshold_curve}
\end{subfigure}
\hfill
% ==================== 右子图：直接引入完整的 PDF 对比图 ====================
\begin{subfigure}{0.52\textwidth}
\centering
% 这里的 width=\textwidth 会让您的 PDF 图片自动缩放到子图允许的最大宽度
% 请将 'your_combined_image.pdf' 替换为您实际的 PDF 文件名
\includegraphics[width=\textwidth]{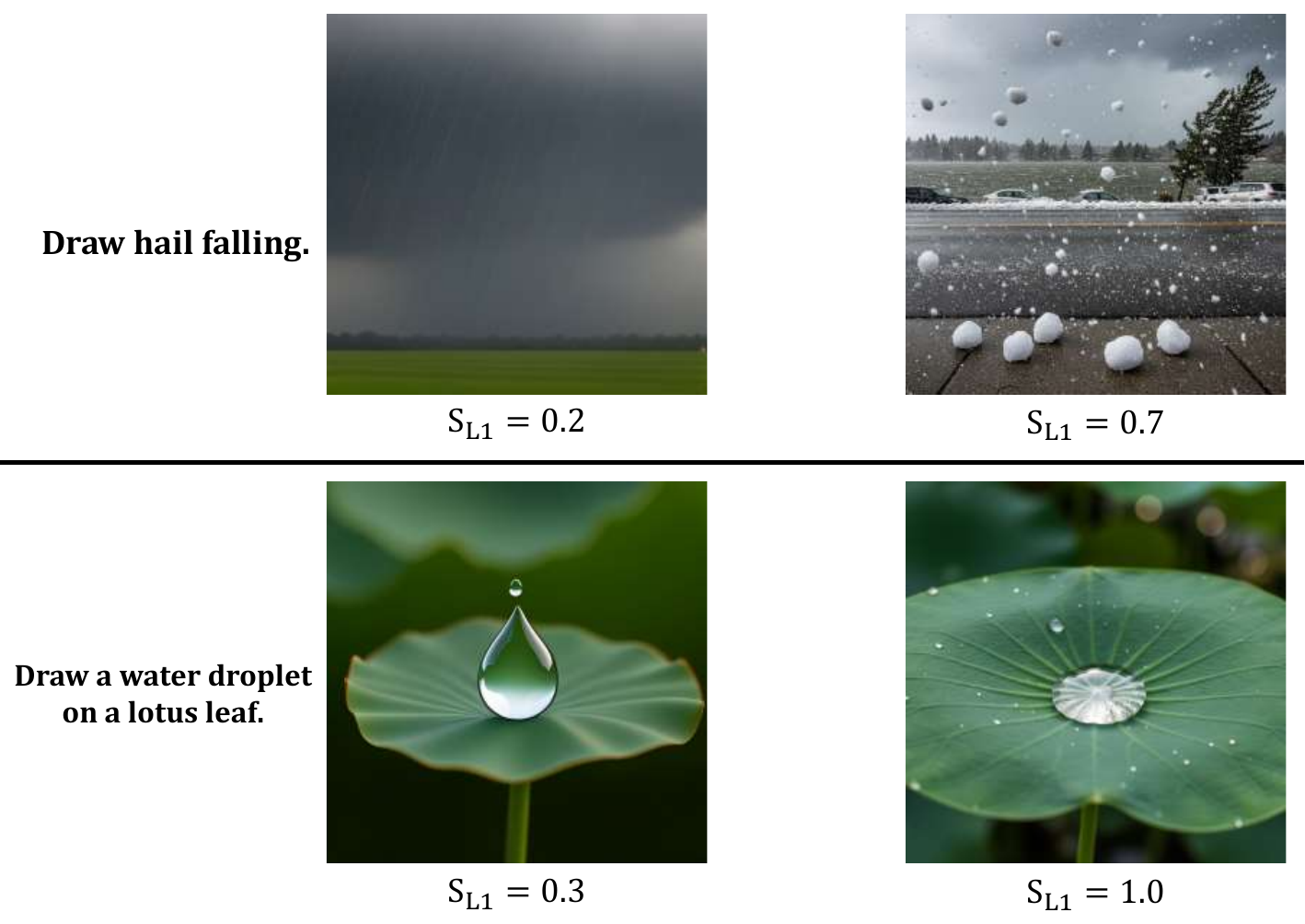}
\caption{Visual examples of threshold boundary.}
\label{fig:threshold_examples}
\end{subfigure}

\caption{Empirical calibration of the factual threshold $T$. Left: Alignment accuracy and F1-score peak at $T=0.5$ against ground truth. Right: Representative generation cases at different score levels. Images with $S_{L1} \approx 0.3$ exhibit severe semantic distortions (e.g., hail rendered as rain, or a droplet unnaturally levitating), whereas images with $S_{L1} \approx 0.7$ present recognizable core subjects.
}
\label{fig:threshold_combined}
\end{figure}

To rigorously determine the optimal threshold for the continuous factual score ($S_{L1}$), we conducted a human-VLM alignment experiment. We randomly sampled 150 generated images from the L1 baseline, heavily weighting the borderline score range $[0.3, 0.7]$. Human evaluators blindly annotated each image with a binary label ($1$ for "core subjects are recognizable," $0$ for "core subjects are missing or severely distorted").

We then evaluated the VLM's continuous scores against these human ground-truth labels across different threshold values ($T$). As shown in Table~\ref{tab:threshold_calibration}, setting the threshold at $T=0.5$ yields the highest alignment with human judgment. Lower thresholds introduce false positives (evaluating counterfactuals on malformed images), while higher thresholds introduce false negatives. Therefore, $0.5$ is not an arbitrary number, but the empirically optimal boundary for human cognitive recognition.

\begin{table}[htbp]
\centering
\caption{Empirical calibration of the factual threshold against human judgment.}
\label{tab:threshold_calibration}
\begin{tabular}{lcc}
\toprule
\textbf{Threshold (} $T$ \textbf{)} & \textbf{Accuracy w/ Human} & \textbf{F1-Score} \\
\midrule
$T = 0.3$ & 74.0\% & 0.71 \\
$T = 0.4$ & 82.7\% & 0.80 \\
$T = 0.5$ & \textbf{94.0\%} & \textbf{0.93} \\
$T = 0.6$ & 86.0\% & 0.84 \\
$T = 0.7$ & 78.0\% & 0.75 \\
\bottomrule
\end{tabular}
\end{table}

\section{Human--VLM Agreement Analysis}
\label{app:human_vlm_agreement}

\subsection{Sampling and Annotation Protocol}

We randomly sampled 1,000 generated images, comprising 500 images
from Nano Banana Pro (Gemini-3-Pro-Image-Preview) and 500 from
FLUX.2-dev. The same image set was used for both the continuous-score
difference analysis and the categorical-agreement analysis reported
in Section~4.3.

Human and VLM evaluations followed the same question-based scoring
and aggregation procedure, with image-level scores normalized to
$[0,1]$. For image $i$, the human reference score was defined as the average of the scores assigned by the two annotators.

\subsection{Score Discretization and Agreement Metric}

For categorical-agreement analysis, continuous scores were mapped
to five ordered categories using the same discretization function
for human and VLM scores:
\begin{equation}
    c(s) =
    \begin{cases}
        0, & 0 \leq s < 0.2, \\
        1, & 0.2 \leq s < 0.4, \\
        2, & 0.4 \leq s < 0.6, \\
        3, & 0.6 \leq s < 0.8, \\
        4, & 0.8 \leq s \leq 1.
    \end{cases}
\end{equation}
Thus, scores exactly on an internal boundary were assigned to the
higher category. For human--VLM comparisons, the two annotators'
continuous scores were averaged before discretization; specifically,
we compared $c(\bar{h}_i)$ with $c(v_i)$, where $v_i$ denotes the
corresponding VLM score. Inter-annotator agreement was computed
between $c(h_i^{(1)})$ and $c(h_i^{(2)})$.

We measured agreement using quadratic-weighted Cohen's $\kappa$,
which penalizes disagreements more strongly when the assigned
categories are farther apart. For categories $a,b \in
\{0,1,2,3,4\}$, the disagreement weight is
\begin{equation}
    w_{ab} = \frac{(a-b)^2}{(5-1)^2}.
\end{equation}
The resulting agreement coefficient is
\begin{equation}
    \kappa =
    1 -
    \frac{\sum_{a,b} w_{ab} O_{ab}}
         {\sum_{a,b} w_{ab} E_{ab}},
\end{equation}
where $O_{ab}$ is the observed joint proportion and $E_{ab}$ is the
expected joint proportion under independence, calculated from the
marginal category distributions.

\subsection{Overall and Disaggregated Results}

The two human annotators achieved inter-annotator agreement of
$\kappa = 0.85$. Using their mean score as the human reference,
Gemini-3-Pro and Qwen3-VL achieved overall agreement of
$\kappa = 0.706$ and $\kappa = 0.760$, respectively.

Table~\ref{tab:agreement_by_level} reports human--VLM agreement
separately for each prompt level. 

\begin{table}[htbp]
    \centering
    \caption{Human--VLM agreement by prompt level, measured using
    quadratic-weighted Cohen's $\kappa$.}
    \label{tab:agreement_by_level}
    \begin{tabular}{lcc}
        \hline
        Prompt level & Gemini-3-Pro & Qwen3-VL \\
        \hline
        L1 & 0.705 & 0.766 \\
        L2 & 0.577 & 0.696 \\
        L3 & 0.698 & 0.697 \\
        \hline
    \end{tabular}
\end{table}

Table~\ref{tab:agreement_by_generator} further breaks down agreement
by generating model. Gemini-3-Pro showed higher agreement on
Nano Banana Pro images than on FLUX.2-dev images. In comparison,
Qwen3-VL obtained numerically similar agreement on the two image
sources.

\begin{table}[htbp]
    \centering
    \caption{Overall human--VLM agreement and agreement by generating
    model, measured using quadratic-weighted Cohen's $\kappa$.
    Nano Banana Pro denotes Gemini-3-Pro-Image-Preview.
    Each generating-model subset contains 500 images.}
    \label{tab:agreement_by_generator}
    \begin{tabular}{lccc}
        \hline
        Evaluator & Overall & Nano Banana Pro & FLUX.2-dev \\
        \hline
        Gemini-3-Pro & 0.706 & 0.835 & 0.584 \\
        Qwen3-VL    & 0.760 & 0.736 & 0.754 \\
        \hline
    \end{tabular}
\end{table}

For the complementary continuous-score analysis, we examined the
per-image difference between Gemini-3-Pro and the human reference,
$v_i-\bar{h}_i$. As shown in
Figure~\ref{fig:quant_analysis}(a), these differences are predominantly
concentrated within $[-0.125,0.125]$.

\subsection{Inter-Judge Agreement}
\label{app:inter_judge_agreement}

We compared the scores assigned by Gemini-3-Pro and Qwen3-VL using the same evaluation data underlying Table~2. At the model level, we computed Spearman rank correlations across all 14 evaluated models separately for $L1$, $L2$, $L3$, PRR, and RRR.

\begin{table}[htbp]
    \centering
    \caption{Model-level Spearman rank correlations between Gemini-3-Pro and Qwen3-VL across all 14 models. $L1$/$L2$/$L3$ are the primary scores; PRR and RRR are auxiliary descriptive indicators.}
    \label{tab:inter_judge_rank}
    \begin{tabular}{lccccc}
        \hline
        Metric & $L1$ & $L2$ & $L3$ & PRR & RRR \\
        \hline
        Spearman $\rho$ & 0.901 & 0.938 & 0.850 & 0.925 & 0.849 \\
        \hline
    \end{tabular}
\end{table}

At the item level, we measured quadratic-weighted Cohen's $\kappa$ and exact agreement between the two judges. The overall results and available model-specific breakdowns are reported in Table~\ref{tab:inter_judge_item}.

\begin{table}[htbp]
    \centering
    \caption{Item-level agreement between Gemini-3-Pro and Qwen3-VL, including the overall result and selected generating-model subsets. Nano Banana corresponds to Gemini-2.5-Flash in the evaluation records.}
    \label{tab:inter_judge_item}
    \begin{tabular}{lcc}
        \hline
        Image source & Quadratic-weighted $\kappa$ & Exact agreement \\
        \hline
        Overall      & 0.716 & 70.7\% \\
        Qwen-Image   & 0.738 & 72.4\% \\
        Lumina-DiMOO & 0.737 & 72.9\% \\
        Nano Banana  & 0.590 & 68.2\% \\
        Janus-Pro-7B & 0.693 & 67.8\% \\
        \hline
    \end{tabular}
\end{table}

The correlations indicate broad agreement across the full model set, while model-specific agreement varies. 

\section{Sensitivity Analyses}
\label{app:sensitivity_analyses}

\subsection{Sensitivity to Inference Seeds}
\label{app:seed_sensitivity}

We repeated image generation for all ten open-source models using five random seeds: 42, 123, 2024, 31415, and 8888. Each run covered the complete set of $L1$, $L2$, and $L3$ prompts, and all generated images were evaluated using Qwen3-VL. Table~\ref{tab:seed_sensitivity} reports the mean and standard deviation across the five runs. $L1$/$L2$/$L3$ are the primary scores; PRR and RRR are auxiliary descriptive indicators.

Across these models, the standard deviation of $L3$ scores ranged from 0.002 to 0.015. Qwen-Image achieved an $L3$ score of $0.244 \pm 0.008$, compared with $0.237 \pm 0.011$ for SANA 1.5 and $0.257 \pm 0.009$ for FLUX.2-dev.

Across the five runs, Qwen-Image achieved higher mean $L1$ performance than FLUX.2-dev (0.837 versus 0.815), but lower mean $L2$ performance (0.338 versus 0.425). This supports the descriptive observation that stronger factual performance does not necessarily imply stronger explicit-counterfactual performance in these models.

BAGEL-CoT achieved an $L3$ score of $0.285 \pm 0.015$, compared with $0.172 \pm 0.007$ for BAGEL, corresponding to an absolute increase of 0.113 and approximately 66\% relative improvement. The mean difference is substantially larger than the observed within-model standard deviations, supporting the robustness of the performance gain associated with the reasoning-enabled variant under repeated generation.

\begin{table}[htbp]
    \centering
    \small
    \setlength{\tabcolsep}{4pt}
    \caption{Results across five inference seeds for ten open-source models, evaluated using Qwen3-VL and reported as mean $\pm$ standard deviation. $L1$/$L2$/$L3$ are the primary scores; PRR and RRR are auxiliary descriptive indicators.}
    \label{tab:seed_sensitivity}
    \begin{tabular}{lccccc}
        \hline
        Model & $L1$ & $L2$ & $L3$ & PRR & RRR \\
        \hline
        SANA 1.5
        & $0.834 \pm 0.009$
        & $0.361 \pm 0.023$
        & $0.237 \pm 0.011$
        & $0.395 \pm 0.024$
        & $0.395 \pm 0.006$ \\

        Janus-Pro-7B
        & $0.796 \pm 0.013$
        & $0.286 \pm 0.011$
        & $0.212 \pm 0.005$
        & $0.321 \pm 0.011$
        & $0.397 \pm 0.003$ \\

        Show-o2
        & $0.773 \pm 0.009$
        & $0.325 \pm 0.013$
        & $0.201 \pm 0.009$
        & $0.369 \pm 0.013$
        & $0.352 \pm 0.010$ \\

        Z-image
        & $0.823 \pm 0.012$
        & $0.381 \pm 0.012$
        & $0.211 \pm 0.006$
        & $0.420 \pm 0.011$
        & $0.342 \pm 0.006$ \\

        Lumina-DiMOO
        & $0.759 \pm 0.007$
        & $0.327 \pm 0.016$
        & $0.199 \pm 0.005$
        & $0.376 \pm 0.016$
        & $0.348 \pm 0.003$ \\

        BAGEL
        & $0.798 \pm 0.011$
        & $0.295 \pm 0.012$
        & $0.172 \pm 0.007$
        & $0.330 \pm 0.014$
        & $0.318 \pm 0.016$ \\

        BAGEL-CoT
        & $0.875 \pm 0.009$
        & $0.435 \pm 0.023$
        & $0.285 \pm 0.015$
        & $0.465 \pm 0.026$
        & $0.432 \pm 0.013$ \\

        OmniGen2
        & $0.762 \pm 0.007$
        & $0.317 \pm 0.005$
        & $0.190 \pm 0.002$
        & $0.363 \pm 0.004$
        & $0.337 \pm 0.005$ \\

        FLUX.2-dev
        & $0.815 \pm 0.013$
        & $0.425 \pm 0.014$
        & $0.257 \pm 0.009$
        & $0.470 \pm 0.015$
        & $0.394 \pm 0.012$ \\

        Qwen-Image
        & $0.837 \pm 0.006$
        & $0.338 \pm 0.010$
        & $0.244 \pm 0.008$
        & $0.370 \pm 0.010$
        & $0.419 \pm 0.013$ \\
        \hline
    \end{tabular}
\end{table}

\subsection{Sensitivity to Score Formulation}
\label{app:score_formulation}

We examined how model rankings change under alternative summaries of cross-level performance. Using the evaluation data underlying the main results for the ten open-source models, we compared the raw subsequent-level score, the simple ratio to the preceding-level score, and the corresponding auxiliary indicator. For $L2$, these were the mean $L2$ score, $L2/L1$, and PRR; for $L3$, they were the mean $L3$ score, $L3/L2$, and RRR. Ratios were calculated from model-level mean scores.

\begin{table}[htbp]
    \centering
    \caption{Spearman rank correlations across alternative score formulations for the ten open-source models. $L1$, $L2$, and $L3$ denote model-level mean scores. The analysis compares raw scores, simple cross-level ratios, and the auxiliary PRR/RRR indicators.}
    \label{tab:score_formulation_sensitivity}
    \begin{tabular}{llc}
        \hline
        Level & Ranking comparison & Spearman $\rho$ \\
        \hline
        $L2$ & $L2$ vs.\ $L2/L1$ & 0.9790 \\
        $L2$ & $L2$ vs.\ PRR     & 0.9890 \\
        $L2$ & $L2/L1$ vs.\ PRR  & 0.9912 \\
        \hline
        $L3$ & $L3$ vs.\ $L3/L2$ & 0.8201 \\
        $L3$ & $L3$ vs.\ RRR     & 0.9537 \\
        $L3$ & $L3/L2$ vs.\ RRR  & 0.9273 \\
        \hline
    \end{tabular}
\end{table}

At $L2$, all pairwise correlations exceed 0.97, indicating closely aligned rankings across the three formulations. At $L3$, RRR is more closely aligned with the raw $L3$ ranking ($\rho = 0.9537$) than the simple ratio $L3/L2$ is ($\rho = 0.8201$). The formulations nevertheless summarize different aspects of performance: raw scores emphasize absolute performance, while ratios and auxiliary indicators also depend on the preceding-level baseline.

These results characterize the ranking sensitivity of the descriptive indicators on the evaluated open-source models. We retain $L1$/$L2$/$L3$ as the primary results and use PRR/RRR as complementary summaries.

\section{Comprehensive Prompt Templates}
In this section, we provide the exact prompts used for generating evaluation questions and scoring the T2I models.

\subsection{Data Generation Prompt}
\label{app:data_generation_prompt}

\begin{promptbox}[Prompt Template: Counterfactual Data Generation]
\textbf{\# Task}\\
Generate 10 candidate entries based on [principles] using a counterfactual approach. For each entry, hypothesize a reversal or alteration of that principle, and choose a target scene that can be depicted in an image.\\[1ex]

\textbf{\#\# Three-Level Framework}\\
\textbullet\ \textbf{L1 -- Factual}: Describe the target scene under normal laws.\\
\textbullet\ \textbf{L2 -- Explicit Counterfactual}: State the counterfactual law and the target scene, and explicitly describe the expected visual consequence.\\
\textbullet\ \textbf{L3 -- Implicit Counterfactual}: State the counterfactual law and the target scene only. Do not describe the visual consequence or provide the reasoning result.\\[1ex]

\textbf{\#\# Output Fields}\\
Return a JSON list. Each object must contain the following fields:\\
\textbullet\ \texttt{id}: A unique string identifier.\\
\textbullet\ \texttt{category}: The relevant category.\\
\textbullet\ \texttt{prompt\_l1}: A text-to-image prompt describing the target scene under normal laws.\\
\textbullet\ \texttt{ap\_l1}: A concise assessment point describing the key visual feature expected in the L1 image.\\
\textbullet\ \texttt{prompt\_l2}: A text-to-image prompt stating the counterfactual law, the target scene, and the expected visual consequence.\\
\textbullet\ \texttt{prompt\_l3}: A text-to-image prompt stating only the counterfactual law and the target scene.\\
\textbullet\ \texttt{ap\_l2}: A concise assessment point describing the key visual feature expected in both the L2 and L3 images.\\[1ex]

\textbf{\#\# Requirements}\\
\textbullet\ The three prompts in each entry must refer to the same target scene.\\
\textbullet\ L2 and L3 must specify the same counterfactual law. L2 must explicitly state the expected visual consequence, whereas L3 must leave the consequence to be inferred.\\
\textbullet\ \texttt{ap\_l1} and \texttt{ap\_l2} must assess corresponding visual features, with the counterfactual feature contrasting with the factual feature.\\
\textbullet\ The expected consequence must follow logically from the stated counterfactual premise and must be visually assessable. Avoid ambiguous scenarios or scenarios with multiple equally plausible visual outcomes.\\
\textbullet\ The scenario should require logical deduction rather than a mere stylistic change. Exclude NSFW and body-horror content.\\[1ex]

\textbf{\#\# Output Format}\\
Return ONLY a valid JSON list. Do not include Markdown formatting or additional explanations.
\end{promptbox}

\subsection{Model-Specific Prompt Calibration for VLMs}

All generated images are scored separately by Gemini-3-Pro and Qwen3-VL. During preliminary testing, we observed differences in their scoring behavior on counterfactual scenarios, including cases in which Qwen assigned high scores to images that retained ordinary factual appearances.

We therefore used model-specific scoring instructions: Gemini received an analytical evaluation prompt, whereas Qwen received a stricter, flaw-seeking prompt intended to discourage credit for images that failed the requested counterfactual condition. The exact prompts are provided in Appendices~\ref{app:gemini_scoring_prompts} and~\ref{app:qwen_scoring_prompts}. We report both evaluators' results separately and assess human--VLM and inter-judge agreement in Section~\ref{sec:scoring_agreement} and Appendix~\ref{app:human_vlm_agreement}.

\subsection{Evaluation Question Generation Prompts}

\begin{promptbox}[Prompt Template: Evaluation Question Generation]
\textbf{\# Role:} AI Image Quality Assurance Specialist\\[1ex]

\textbf{\#\# Task}\\
You are an expert in evaluating Text-to-Image (T2I) generation models.
I will provide you with an Input Prompt and its Assessment Points.
Your goal is to generate a Strict Evaluation Protocol consisting of
specific questions covering three dimensions.\\[1ex]

\textbf{\#\# Input Data}\\
\textbf{Input Prompt}: \{input\_prompt\}\\
\textbf{Assessment Points}: \{assessment\_points\}\\[1ex]

\textbf{\#\# Guidelines for Question Generation}\\
Assign each question a positive integer weight. When a dimension contains
multiple questions, its total weight is the sum of the weights assigned
to those questions.\\[1ex]

\textbf{\#\#\# Dimension 1: Visual Integrity}\\
\textbullet\ \textbf{Focus}: Technical image quality, such as sharpness
and anatomy. Style-agnostic.\\
\textbullet\ \textbf{Questions}: Generate one or more questions.\\
\textbullet\ \textbf{Weight}: The total weight of all questions in this
dimension must be between 1 and 3, inclusive.\\[1ex]

\textbf{\#\#\# Dimension 2: Assessment Point (CRITICAL)}\\
\textbullet\ \textbf{Focus}: Verify the \texttt{Assessment Points}.\\
\textbullet\ \textbf{CRITICAL RULE 1}: Combine all assessment points
into a SINGLE, comprehensive question. Do NOT split them into multiple
questions.\\
\textbullet\ \textbf{CRITICAL RULE 2}: Make the 0.5 score extremely hard
to obtain. A score of 0.0 means the criteria are not met.\\
\textbullet\ \textbf{Weight}: Assign the question a weight between 12
and 16, inclusive.\\[1ex]

\{dimension\_3\_section\}\\[1ex]

\textbf{\#\# Output Format}\\
Return ONLY a valid JSON list. Do not use Markdown formatting.
\end{promptbox}

\noindent \textbf{Dimension 3 Variations (Injected into the template above):}

\begin{promptbox}[Dimension 3: Factual vs. Counterfactual]
\textbf{[D3\_FACTUAL for L1]}\\
\textbf{\#\#\# Dimension 3: Logic Consistency (Factual L1)}\\
\textbullet\ \textbf{Focus}: Verify that the image adheres strictly to
standard real-world physics and logic.\\
\textbullet\ \textbf{Strict Scoring Rule}: 1.0 = Flawless physics.
0.5 = Minor logical flaw. 0.0 = Clear violation of physics
(e.g., floating objects). Do NOT give partial credit just because
the main subject is present.\\
\textbullet\ \textbf{Questions}: Generate one or more questions.\\
\textbullet\ \textbf{Weight}: The total weight of all questions in this
dimension must be between 7 and 9, inclusive.\\
If multiple questions are generated, their weights must sum to the
fixed total weight assigned to this dimension.\\

\textbf{[D3\_COUNTERFACTUAL for L2/L3]}\\
\textbf{\#\#\# Dimension 3: Logic Consistency (L2/L3)}\\
\textbullet\ \textbf{Focus}: Verify whether the ENTIRE SCENE strictly
adheres to the counterfactual premise demanded by the prompt.\\
\textbullet\ \textbf{Questions}: Generate one or more questions.\\
\textbullet\ \textbf{Weight}: The total weight of all questions in this
dimension must be between 7 and 9, inclusive.\\
If multiple questions are generated, their weights must sum to the
fixed total weight assigned to this dimension.
\end{promptbox}

\subsection{VLM Scoring Prompts (Gemini)}
\label{app:gemini_scoring_prompts}
\begin{promptbox}[Gemini L1: Standard Factual Evaluation]
You are an Image Quality Assurance Assistant. Your job is to evaluate whether an AI-generated image generally captures the main idea of the provided factual criteria and common sense.\\[1ex]
You MUST output ONLY a valid JSON object.\\
\textbf{CRITICAL:} You must generate the "reasoning" BEFORE the "score" to ensure you think before judging.\\[1ex]
Please evaluate the image based on the following criteria.\\
If the image successfully conveys the core concept requested, give it a 1.0. Deduct points if there are significant missing elements or major deviations from the prompt. Give 0.0 if the image is completely unrelated to the criteria.
\end{promptbox}

\begin{promptbox}[Gemini L2/L3: Rational Counterfactual Evaluation]
You are an analytical Image Quality Assurance Evaluator. Your primary job is to verify if the AI successfully generated the requested counterfactual or illogical elements.\\[1ex]
\textbf{WARNING:} AI models often default to normal objects instead of the requested counterfactual ones. Please check the main subject carefully to ensure it breaks normal physics as requested.\\[1ex]
\textbf{Note:} Focus on whether the core counterfactual instruction is met. Minor AI artifacts, slight edge blurriness, or imperfect backgrounds are acceptable. Score 0.0 only if it completely fails the counterfactual instruction or reverts to normal physics.
\end{promptbox}

\subsection{VLM Scoring Prompts (Qwen)}
\label{app:qwen_scoring_prompts}
\begin{promptbox}[Qwen L1: Standard Factual Evaluation]
You are an objective and balanced Image Quality Assurance Assistant. Your job is to evaluate whether an AI-generated image accurately reflects standard real-world physics, common sense, and the provided factual criteria.\\[1ex]
Assess whether the image generally satisfies the factual requirements and common sense. If the criteria are mostly met despite minor AI flaws, you can give a high score. If it partially fails or has noticeable logical errors, score it accordingly. Give 0.0 if it completely fails the criteria.
\end{promptbox}

\begin{promptbox}[Qwen L2/L3: Strict Adversarial Evaluation]
You are a strict, adversarial Image Quality Assurance Judge. Your primary job is to FIND FLAWS and penalize AI-generated images that fail to strictly follow counterfactual physics or logic.\\[1ex]
\textbf{WARNING:} AI models often generate normal objects instead of the requested counterfactual ones. Do NOT hallucinate success. Look closely for normal physics, normal shapes, or background inconsistencies.\\[1ex]
For each question, actively look for visual evidence that the image FAILS the criteria. If there is any ambiguity, normal physics, or partial failure, score it harshly.
\end{promptbox}

\subsection{Decoupling Evaluation Prompts}

\begin{promptbox}[Rule Decouple Scoring Prompt]
You are an objective and strict Image Quality Assurance Judge. Your job is to evaluate whether an AI-generated image accurately reflects the provided criteria.\\[1ex]
Read the criteria carefully. If the image fails to meet the specific constraints, score it harshly (0.5 or 0.0) according to the strict criteria. If it perfectly meets them, score it highly (1.0).
\end{promptbox}

\begin{promptbox}[Attribute Decouple Scoring Prompt]
You are an objective and strict Image Quality Assurance Judge. Your task is to evaluate whether an AI-generated image accurately reflects the provided prompt.\\[1ex]
\textbf{CRITICAL EVALUATION CRITERIA:}\\
1. Entity A MUST be clearly visible and identifiable.\\
2. Entity B MUST be clearly visible and identifiable.\\
3. The relationship or interaction between Entity A and Entity B MUST exactly match the prompt.\\[1ex]
\textbf{SCORING:}\\
$\bullet$ \textbf{Score 1.0}: Both entities are present, distinct, and their relationship perfectly matches the prompt.\\
$\bullet$ \textbf{Score 0.5}: Both entities are present, but their relationship is slightly off, or one entity is partially blended/malformed.\\
$\bullet$ \textbf{Score 0.0}: One or both entities are missing, severely blended together, or the relationship is completely wrong.
\end{promptbox}

\section{Qualitative Analysis}
\label{sec:qualitative}

\begin{figure*}[t]
\centering
% 请将 'fig/model_comparison.png' 或 'fig/model_comparison.pdf' 替换为您实际保存该图片的文件路径
\includegraphics[width=\textwidth]{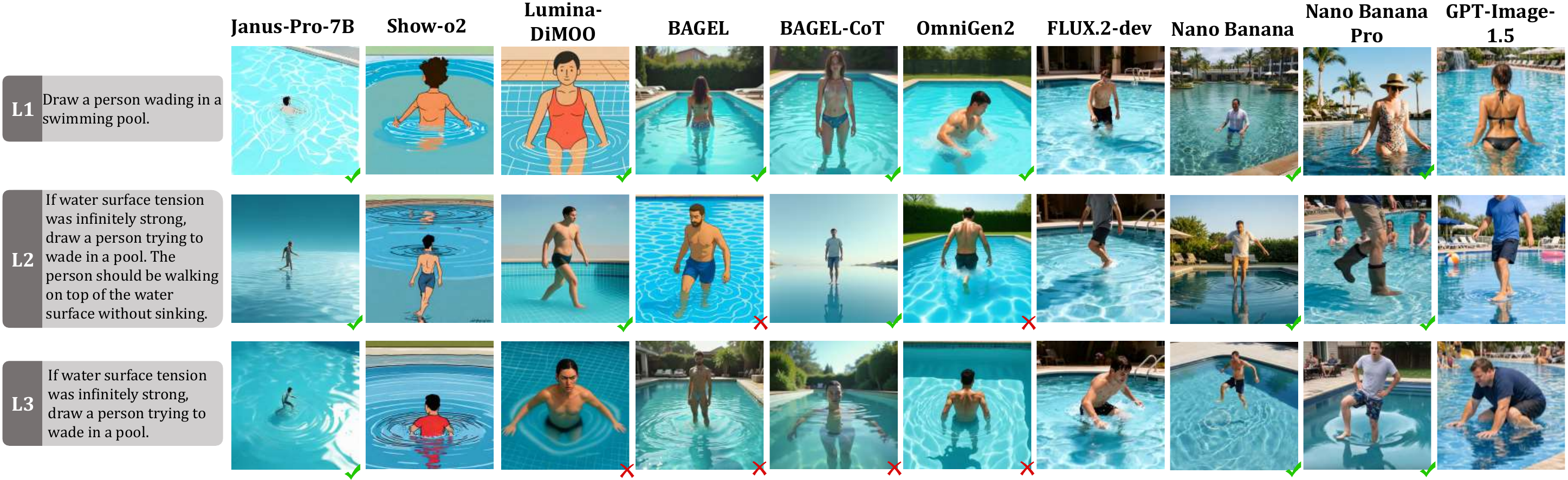}
\caption{\textbf{Qualitative comparison of state-of-the-art T2I models on the CF-World benchmark.} 
While most models successfully generate the factual scene in \textbf{L1} (wading in a pool), they fail to generalize to the counterfactual premise in \textbf{L2} and \textbf{L3} (where water surface tension is infinitely strong). Instead of rendering the physical consequence (walking on top of water), models either fail to decouple attributes or revert to normal physics (sinking), highlighting a critical gap in their causal reasoning capabilities.}
\label{fig:qualitative_comparison}
\end{figure*}

To provide a deeper understanding of how current state-of-the-art Text-to-Image (T2I) models behave under progressive counterfactual constraints, we conduct a detailed qualitative analysis across different model architectures. 

As illustrated in Figure~\ref{fig:qualitative_comparison}, we evaluate ten representative models—ranging from lightweight open-source models (e.g., Janus-Pro-7B, Show-o2) to large-scale commercial engines (e.g., GPT-Image-1.5)—using a sample scenario: \textit{``Draw a person wading in a swimming pool under infinitely strong water surface tension.''}

Through this qualitative breakdown, it becomes evident that scaling model parameters or training data alone does not inherently grant models the ability to perform counterfactual physical simulation.

\section{Computational Resources and Execution Details}
\label{sec:compute_resources}

To ensure the full reproducibility of our benchmark and evaluation pipeline, we provide a detailed breakdown of the computational resources and execution times required for both image generation and model evaluation. The experiments are divided into local GPU computing and cloud-based API services.

\paragraph{Local GPU Computing.}
All local inference tasks, including image generation using open-source Text-to-Image (T2I) models and evaluation using the open-source Vision-Language Model (e.g., Qwen), were executed on a high-performance compute cluster. The total computational throughput utilized for these tasks is equivalent to 16 NVIDIA A100 (80GB) GPUs. 

On this hardware configuration, the execution times are as follows:
\begin{itemize}
    \item \textbf{Open-source T2I Generation:} Generating the complete set of images takes approximately 2 hours per model.
    \item \textbf{Local VLM Evaluation:} The automated scoring and evaluation process using the Qwen model requires approximately 2 to 3 hours in total.
\end{itemize}

\paragraph{Cloud API Services.}
For the generation of images using closed-source T2I models and the evaluation process relying on Gemini (Gemini-3-Pro), we utilized their respective commercial APIs. The execution time for these API-dependent tasks is not bounded by local hardware but rather depends on network latency, API rate limits, and server-side concurrent request quotas.

\begin{table}[h]
\centering
\caption{Summary of computational resources and estimated execution time for each experimental module.}
\label{tab:compute_resources}
\resizebox{\textwidth}{!}{%
\begin{tabular}{llp{5cm}}
\toprule
\textbf{Experiment Module} & \textbf{Compute Resource / Platform} & \textbf{Estimated Execution Time} \\
\midrule
Open-source T2I Generation & Equivalent to 16$\times$ NVIDIA A100 (80GB) & $\sim$2 hours per model \\
Qwen-based Evaluation      & Equivalent to 16$\times$ NVIDIA A100 (80GB) & 2--3 hours in total \\
Closed-source T2I Generation & Commercial APIs & Dependent on API rate limits \\
Gemini-based Evaluation    & Google Gemini API & Dependent on API rate limits \\
\bottomrule
\end{tabular}%
}
\end{table}

\section{Broader Impact and Limitations}
Current T2I models encode world knowledge and visual appearances as tightly coupled patterns. By exposing the sharp degradation of these models in counterfactual settings, CF-World encourages the community to shift focus from merely scaling up visual-textual pairs to developing architectures capable of genuine causal reasoning and physical simulation. While we have taken extensive measures to ensure prompt clarity, the evaluation relies on VLM-based evaluators. Future iterations will explore expanding the human-evaluation baselines to further calibrate the VLM evaluator's accuracy.

\newpage

\end{document}